\documentclass[10pt,twocolumn,letterpaper]{article}

\usepackage{cvpr}              
\definecolor{cvprblue}{rgb}{0.21,0.49,0.74}
\usepackage[pagebackref,breaklinks,colorlinks,allcolors=cvprblue]{hyperref}

\def\paperID{} 
\def\confName{CVPR}
\def\confYear{2026}

\usepackage{graphicx} 
\usepackage{booktabs}
\usepackage{multirow}
\usepackage[table]{xcolor} 
\usepackage{amssymb}  
\usepackage{diagbox}
\usepackage{makecell}
\usepackage{subcaption}     
\usepackage{adjustbox}

\usepackage{algorithm}
\usepackage{algorithmic}
\usepackage{amsmath} 

\usepackage{appendix} 
\usepackage{minitoc}  

\usepackage{bbding}

\title{TTL: Test-time Textual Learning for OOD Detection with Pretrained Vision-Language Models}

\author{
Jinlun Ye$^{1,3,5}$, Jiang Liao$^{2}$, Runhe Lai$^{1,3,5}$, Xinhua Lu$^{1,3,5}$, 
Jiaxin Zhuang$^{4}$\thanks{Corresponding author}, Zhiyong Gan$^{2}$, Ruixuan Wang$^{1,3,5}$\footnotemark[1]\\
$^{1}$Sun Yat-sen University \quad
$^{2}$China United Network Communications Corporation Limited Guangdong Branch \\
$^{3}$Peng Cheng Laboratory \quad
$^{4}$Hong Kong University of Science and Technology \\
$^{5}$Key Laboratory of Machine Intelligence and Advanced Computing, MOE \\
{\tt\small jzhuangad@connect.ust.hk, wangruix5@mail.sysu.edu.cn} 
}

\begin{document}
\maketitle
\begin{abstract}
Vision-language models (VLMs) such as CLIP exhibit strong Out-of-distribution (OOD) detection capabilities by aligning visual and textual representations. 
Recent CLIP-based test-time adaptation methods further improve detection performance by incorporating external OOD labels.
However, such labels are finite and fixed, while the real OOD semantic space is inherently open-ended. Consequently, fixed labels fail to represent the diverse and evolving OOD semantics encountered in test streams.
To address this limitation, we introduce \textbf{T}est-time \textbf{T}extual \textbf{L}earning (TTL), a framework that dynamically learns OOD textual semantics from unlabeled test streams, without relying on external OOD labels.
TTL updates learnable prompts using pseudo-labeled test samples to capture emerging OOD knowledge. To suppress noise introduced by pseudo-labels, we introduce an OOD knowledge purification strategy that selects reliable OOD samples for adaptation while suppressing noise.
In addition, TTL maintains an OOD Textual Knowledge Bank that stores high-quality textual features, providing stable score calibration across batches.
Extensive experiments on two standard benchmarks with nine OOD datasets demonstrate that TTL consistently achieves state-of-the-art performance, highlighting the value of textual adaptation for robust test-time OOD detection. 
Our code is available at \href{https://github.com/figec/TTL}{https://github.com/figec/TTL}.
\end{abstract}

\section{Introduction}

\begin{figure}[ht]
    \centering
    \includegraphics[width=0.48\textwidth]{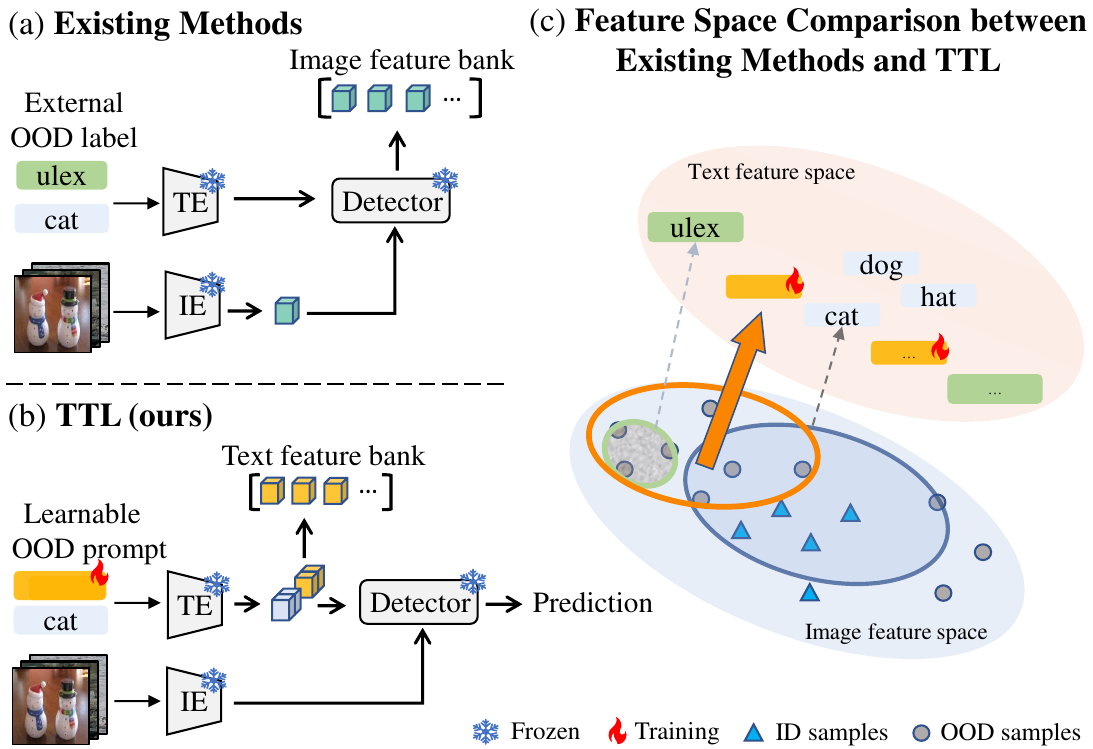}
    
    \vspace{-4pt}
    \caption{Comparison with existing OOD adaptation methods. (a) Existing methods adapt visual features within a fixed text space, limiting adaptation to the samples that fall outside these predefined semantic scopes. (b) Our proposed TTL framework performs test-time textual adaptation by learning and purifying new text knowledge from unlabeled test streams. By aligning these learnable OOD prompts with OOD pseudo-labeled images, it captures clearer OOD knowledge across baches and enhances OOD detection performance. (c) Feature space visualization showing how TTL's adaptive text representations better align with image features compared to the existing methods.
    }
    \label{fig:comparison}  
\end{figure}

Deep learning demonstrates remarkable performance in closed-set scenarios where training and testing data follow identical distributions. However, when deployed in open-world environments, models inevitably encounter OOD data from unknown classes. Critically, models often misclassify such OOD samples as high-confidence in-distribution (ID) classes~\citep{DBLP:conf/cvpr/NguyenYC15, DBLP:conf/iclr/HendrycksG17}, posing severe safety risks in critical applications such as autonomous driving and medical diagnosis. Therefore, accurately detecting OOD data is essential to ensure the reliability and safety of AI systems in real-world deployments.

Traditional OOD detection methods~\citep{DBLP:conf/iclr/HendrycksG17, DBLP:conf/icml/HendrycksBMZKMS22, DBLP:conf/nips/LiuWOL20, DBLP:conf/nips/SunGL21} rely on pre-trained ID classifiers but are limited to the visual modality. The emergence of VLMs like CLIP~\citep{clip} has enabled leveraging multi-modal information for enhanced OOD detection. Recent VLM-based methods have been dedicated to learn OOD knowledge through external images or text labels~\citep{wang2023clipn, neglabel}, or extract such knowledge solely from ID training data, such as background regions~\citep{locoop} or randomly cropped images~\citep{zeng2025local}. However, OOD features derived from specific datasets inherently fail to capture the infinite OOD distributions encountered in the real world.


In order to obtain more practical OOD knowledge, recent methods employ test-time adaptation to adapt VLMs to true OOD distributions. Some intuitive methods \citep{cao2025zsntta} train the OOD detector using pseudo-labeled test samples. 
However, these batch-wise parameter update methods can easily lead to catastrophic forgetting and unstable OOD detection performance~\citep{oodd}.
To mitigate catastrophic forgetting and detection fluctuation, while OODD~\citep{oodd} only store visual features during testing,
AdaNeg~\citep{zhang2024adaneg} additionally leverages the textual modality by aligning external OOD text semantics with actual test distributions, achieving stronger OOD detection performance. 
Nevertheless, AdaNeg depends on a finite and fixed set of OOD labels to represent the open-ended OOD semantic space, which is inherently insufficient. As a result, it struggles to adapt to OOD samples that fall outside these predefined semantic scopes (as shown in Figure~\ref{fig:comparison}c), leading to limited performance. Inspired by prompt learning~\citep{coop}, fine-tuning prompts enables text features to better align with actual data distributions. We thus pose a natural question:

\vspace{4pt} 
\noindent\textit{Would directly learning OOD textual semantics from the test stream—rather than aligning specific labels with OOD distributions—yield better adaptation results?}
\vspace{4pt} 

To this end, we propose \textbf{T}est-time \textbf{T}extual \textbf{L}earning (TTL), a novel framework that dynamically learns OOD textual semantics from the unlabeled test streams and eliminates the dependence on external OOD labels.
%
Our key insight lies in learning a set of OOD prompts that contain diverse and clearer OOD knowledge, ultimately improving OOD detection performance.
%
%
Specifically, we introduce a learnable OOD prompt for each ID class.
Our TTL can acquire valuable OOD textual semantics by amplifying the semantic similarity between the test samples with assigned OOD pseudo-labels and the learnable OOD prompts.
%
Moreover, we propose an OOD knowledge purification strategy to suppress noise from pseudo-labels by decreasing the semantic similarity between OOD prompts and ID boundary samples (\textit{i.e}., low-confidence pseudo-OOD samples).
%
In this way, the OOD prompts can learn clearer OOD knowledge across test streams and facilitate advanced discernment between ID and OOD data.
%
Furthermore, to ensure stable detection and provide broader semantic coverage through diverse distributions, TTL maintains an OOD textual knowledge bank that maintains a dynamic repository of high-quality OOD text features.
%
During inference, the OOD textual knowledge bank is used to calibrate
the final OOD detection score.  
Our key contributions can be summarized as follows:


\begin{itemize}
    \item We propose a Test-time Textual Learning (TTL) framework that dynamically learns OOD textual semantics from test streams without relying on external OOD labels.

    \item We present a novel purification strategy to reduce pseudo-label noise and obtain clearer OOD knowledge.
    
    \item Comprehensive experiments across multiple benchmarks demonstrate that TTL achieves consistent and significant improvements over state-of-the-art methods, including average gains of 12.67\% FPR95 and 3.94\% AUROC.


\end{itemize}

\section{Related Work}

\noindent\textbf{OOD Detection.}
Traditional OOD detection methods focus on single-modal image analysis, falling into two categories. The first designs scoring functions using model outputs (e.g., logits, features, layer statistics)~\citep{DBLP:conf/iclr/HendrycksG17, odin, DBLP:conf/nips/LiuWOL20, DBLP:conf/nips/SunGL21}. The second explores ID-OOD decision boundaries via various training strategies~\citep{vos, DBLP:conf/iclr/MingSD023, tagfog, fodfom}. Though achieving satisfactory results, they overlook textual modalities' rich semantic information, leading to sub-optimal performance~\citep{neglabel}.

To leverage textual knowledge, recent research has focused on employing vision-language models like CLIP~\citep{clip} with powerful multimodal understanding capabilities. These VLM-based methods can be categorized into three main strategies. Concept matching methods such as MCM~\citep{mcm} and CMA~\citep{cma} leverage CLIP's image-text alignment to generate OOD scores based on category names or auxiliary concepts. GL-MCM~\citep{gl-mcm} extends MCM to multi-object scenarios. ID-enhanced methods improve discrimination by exploiting additional information from ID data. For example, FA~\citep{fa} uses ID prompts as references for learnable prompt optimization; LoCoOp~\citep{locoop} and SCT~\citep{sct} employ entropy maximization to reduce background sensitivity; 
OSPCoOp~\citep{ospcoop}, IDLike~\citep{idlike}, Negprompt~\citep{negprompt}, and Local-Prompt~\citep{zeng2025local} generate pseudo-OOD samples through background extraction, image cropping, or assuming the relationship between OOD distribution and ID distribution. 
External knowledge-based methods improve OOD detection using OOD-related information. For example,  Neglabel~\citep{neglabel}, CSP~\citep{csp}, and NegRefine~\citep{negrefine} collect potential OOD labels from large-scale corpora, while CLIPN~\citep{wang2023clipn} learns negative prompts from massive datasets. APT~\citep{APT} employs entropy maximization by introducing external OOD data. However, such methods prove impractical due to the inherent diversity and infinite nature of real-world OOD distribution.
In order to obtain more practical OOD knowledge, 
there has been growing interest in leveraging information from real-time testing scenarios to assist OOD detection~\citep{dcac}. AUTO~\citep{auto} updates the parameters of all batch normalization layers and the final feature block in the model by 
reducing the prediction confidence of potential OOD samples. Unlike updating the original model, AdaND~\citep{cao2025zsntta}
trains an additional noise detector. 
On the other hand,
OODD~\citep{oodd} maintains a priority queue to accumulate more representative OOD image features, which are used to calibrate detector outputs for test samples. AdaNeg~\citep{zhang2024adaneg} further exploits external textual labels to guide the selection and storage of visual features, achieving stronger performance. Despite these advances, existing test-time methods focus primarily on visual-side adaptation with fixed text labels.
In contrast,  our method actively learns discriminative OOD textual knowledge during testing, directly exploiting the adaptive potential of the text modality.

\noindent\textbf{Prompt Learning.} Originated in NLP as a replacement for manual prompt engineering, prompt learning has been adapted to vision-language scenarios with VLMs like CLIP \citep{clip} serving as strong baselines. Methods such as CoOp~\citep{coop} use learnable vectors for template words, boosting CLIP's performance across tasks. While applied to OOD detection~\citep{idlike, fa, zeng2025local}, existing methods operate in training-time with labeled ID data. Ours is the first to apply it to test-time OOD detection, adaptively learning textual knowledge aligned with real-world OOD distributions—enabling dynamic adaptation without pre-defined OOD categories or external datasets.
\section{Methodology}
\subsection{Preliminaries}
\begin{figure*}[h]
\begin{center}
\includegraphics[width=\textwidth]{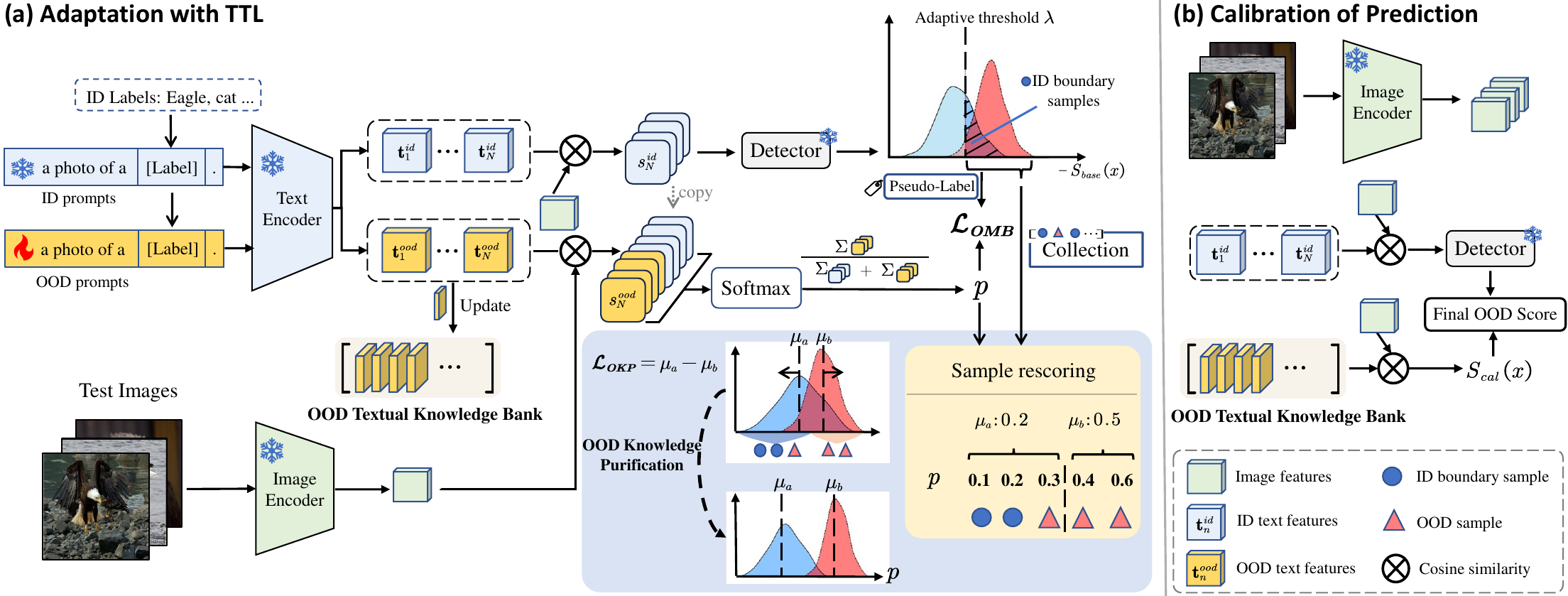}
\end{center}
\vspace{-4pt}
\caption{Overview of the proposed TTL framework. (a) \textbf{Adaptation with TTL:} 
During test time adaptation, pseudo labels produced by a base OOD detector are used to optimize the learnable OOD prompts, allowing the model to gradually acquire OOD textual knowledge.
To reduce noise in pseudo labels, an OOD knowledge purification strategy is introduced to distinguish reliable OOD samples from ID boundary samples. 
The learned OOD textual features are then updated in the OOD textual knowledge bank. (b) \textbf{Calibration of Prediction:} During inference, the base detector’s predictions are further calibrated using the OOD textual knowledge bank.
} 
\label{fig:framework}
\end{figure*}


\noindent\textbf{OOD Detection.} 
The goal of OOD detection is to distinguish ID samples from OOD ones. For a test sample~\(\mathbf{x}\), this can be formulated as a binary classification task.
The decision function $D(\mathbf{x})$ can be defined as
\begin{equation}
D(\mathbf{x}) = 
\begin{cases} 
1~(ID), & \text{if } S(\mathbf{x}) \geq \lambda \\
0~(OOD), & \text{if } S(\mathbf{x}) < \lambda  
\end{cases} \,,
\label{eq:ood_detection}
\end{equation}
where the score function $S(\cdot)$ is obtained from a base OOD detector, and $\lambda$ is the decision threshold.

\noindent\textbf{CLIP and Threshold Determination.} 
CLIP-based OOD detection leverages both visual and textual modalities for improved discrimination. 
CLIP contains two encoders.
The image encoder $f(\cdot)$ converts an input image $\mathbf{x}$ into a feature vector $\mathbf{z} = f(\mathbf{x}) \in \mathbb{R}^d$.
The text encoder $g(\cdot)$ converts the manual template $\mathbf{u}_c$  (\textit{i.e.} ``a photo of a \{\textit{class label}\}") into a feature vector $\mathbf{t}_c = g(\mathbf{u}_c) \in \mathbb{R}^d,c\in\{1,\cdots,N\}$, where $N$ represents the number of ID classes.
We refer to these templates $\mathbf{u}_c$ as ID prompts.
The cosine similarity $\cos(\mathbf{z}, \mathbf{t}_c)$ measures the degree of image-text alignment. 
For CLIP-based OOD detection, following prior study~\cite{DBLP:conf/iccv/Li0SJ23}, the decision threshold $\lambda$ is adaptively determined by minimizing intra-class variance based on the bimodal distribution of OOD detection scores as follows,
{\small
\begin{equation}
\label{eq:adaptive_threshold}
\min_\lambda \frac{1}{N_\text{id}} \sum_{S(\mathbf{x}_i)>\lambda} {(S(\mathbf{x}_i)-\mu_\text{id})^2} +
\frac{1}{N_\text{ood}}\sum_{S(\mathbf{x}_j)\leq\lambda}{(S(\mathbf{x}_j)-\mu_\text{ood})^2},
\end{equation}
}
where $\mu_\text{id} = \frac{1}{N_\text{id}}\sum_{S(\mathbf{x}_i)>\lambda} S(\mathbf{x}_i)$ and $\mu_\text{ood} = \frac{1}{N_\text{ood}}\sum_{S(\mathbf{x}_j)\leq\lambda} S(\mathbf{x}_j)$ are the mean scores above and below $\lambda$, respectively. $N_\text{id}$ and $N_\text{ood}$ denote the number of samples above and below $\lambda$, respectively.
Then the decision threshold $\lambda$ can be applied to Eq.~(\ref{eq:ood_detection}) to obtain the pseudo-label $\hat{y}$ for each test sample $\mathbf{x}$ during test-time.


\subsection{Overview}

In this work, we propose \textbf{T}est-time \textbf{T}extual \textbf{L}earning (TTL), a framework that adaptively learn OOD textual semantics directly from unlabeled test streams. Rather than relying on predefined OOD labels, TTL incrementally learns OOD textual knowledge that reflects the evolving test distribution.
As shown in Figure~\ref{fig:framework}, the adaptation with TTL consists of three components. (1) \textit{OOD Knowledge Learning},
which optimizes learnable OOD prompts using pseudo-labeled test samples to extract OOD textual knowledge;
(2) \textit{OOD Knowledge Purification}, which reduces pseudo-label noise by distinguishing reliable OOD samples from ID boundary samples;
and (3) \textit{OOD Textual Knowledge Bank}, which maintains a dynamic repository of learned OOD text features.
During inference, the bank is used for robust score calibration across test batches. 

\subsection{Test-time Textual Learning}
\label{sec:TTTL}

\noindent\textbf{OOD Knowledge Learning.} 
Our objective is to learn textual representations that dynamically capture discriminative OOD semantics during test-time, instead of relying on external fixed OOD labels. 
%
In order to learn OOD textual semantics directly from the unlabeled test streams, we introduce $N$ learnable OOD prompts $\{\mathbf{u}^{\text{ood}}_i\}_{i=1}^{N}$, each corresponding to an ID class.
Specifically, each learnable OOD prompt is initialized using the same manual template as the ID prompt, to effectively utilize the rich semantic information provided by CLIP’s prior knowledge~\cite{coop}.
The class labels are utilized in OOD prompts to provide foundational semantic information associated with each class name.
In particular, we freeze the ID prompts, the class label part of OOD prompts, and the image and text encoders of CLIP to maintain its generalization capability, while only making the prefix (\textit{i.e.} ``a photo of a'') of OOD prompt learnable.
Based on the OOD prompts, we can obtain the OOD probability $p(\mathbf{x})$ for each test sample $\mathbf{x}$ as follows,
\vspace{-0.5em}
\begin{equation}
     p(\mathbf{x}) =  \frac{\sum_{k=1}^Ns(\mathbf{x}, \mathbf{t}_k^{\text{ood}})}{\sum_{j=1}^{N} s(\mathbf{x}, \mathbf{t}_j^{\text{id}}) + \sum_{j=1}^{N} s(\mathbf{x}, \mathbf{t}_j^{\text{ood}})},
\label{eq:ood_prob}
\end{equation}
where $\mathbf{t}^{\text{id}}$ and $\mathbf{t}^{\text{ood}}$ are the text features for the ID and OOD prompts, respectively, \(s(\mathbf{x},\mathbf{t}) = \exp\left(\cos(f(\mathbf{x}), \mathbf{t}) / \tau\right)\),  and $\tau$ is the temperature hyperparameter.
Then, the learnable OOD prompts can be optimized by our proposed OOD-focused minority-balanced loss $\mathcal{L}_{\text{OMB}}$ as follows,
\vspace{-0.5em}
\begin{equation}
\begin{aligned}
\label{eq:cbce_loss}
 \mathcal{L}_{\text{OMB}} =
 -\frac{1}{\pi_{+}} \sum_{i:\hat{y}_i = 1} \log(1 - p(\mathbf{x}_i))
 -\frac{1}{\pi_{-}} \sum_{j:\hat{y}_j = 0} \log p(\mathbf{x}_j),
\end{aligned}
\end{equation}
where $\pi_{+}$ and $\pi_{-}$ denote the proportions of ID and OOD samples labeled by pseudo-labels $\hat{y}$, respectively.
Specially, $\pi_{+}$ and $\pi_{-}$ are used to mitigate the imbalanced ID and OOD composition in the test stream.
Based on this $\mathcal{L}_{\text{OMB}}$, OOD knowledge can be learned from test streams by amplifying the cosine similarity between the image features of test samples with OOD pseudo-labels and the text features of learnable OOD prompts.

\noindent\textbf{OOD Knowledge Purification.}
Although pseudo-labeled supervision enables valuable OOD knowledge to be learned, pseudo-label noise is unavoidable. As a result, the pseudo-labeled OOD set inevitably contains misclassified ID samples, referred to as ID boundary samples. These boundary samples bias the learned OOD prompts toward ID semantics, weakening their discriminative ability.
However, existing test-time adaptation methods~\citep{cao2025zsntta} for OOD detection typically update features or parameters based on base detector results, without explicitly handling such noise. 
%
Moreover, as adaptation proceeds over batches, the influence of ID boundary samples will be gradually amplified.

To mitigate this issue, we propose an OOD Knowledge Purification (OKP) strategy to reduce the noise induced by ID boundary samples.
%
%
Specifically, 
we first collect a set of pseudo-labeled OOD samples in each test batch.
%
In order to partition this set
into subsets of high-confidence and low-confidence pseudo-OOD samples, we utilize the corresponding OOD probabilities as scores in Eq.~(\ref{eq:adaptive_threshold}) to derive an adaptive threshold $\theta$.
The samples in the set with OOD probabilities above $\theta$ is designated as $S_h=\left\{i\mid p(\mathbf{x}_i)>\theta\right\}$, and the remaining samples are designated as $S_{\ell}=\left\{j\mid p(\mathbf{x}_j)\leq\theta\right\}$, with $S_{\ell}$ containing ID boundary samples.
Based on $S_h$ and $S_{\ell}$, we propose the OOD knowledge purification loss $\mathcal{L}_{\text{OKP}}$ as follows,
\begin{equation}
\label{eq:ditance_loss}
\mathcal{L}_{\text{OKP}} = -\left( \frac{1}{|S_h|} \sum_{i \in S_h} p(\mathbf{x}_i) - \frac{1}{|S_{\ell}|} \sum_{j \in S_{\ell}} p(\mathbf{x}_j) \right) \,.
\end{equation}
The proposed $\mathcal{L}_{\text{OKP}}$ can effectively suppress the interference from ID boundary samples during OOD prompt optimization across test streams.
%
This is because the semantic similarity between ID boundary samples and the OOD prompt decreases, while that between high-confidence OOD samples and the OOD prompt further increases.
%
Hence, our $\mathcal{L}_{\text{OKP}}$ enables the learnable OOD prompt to acquire clearer OOD knowledge and facilitates advanced discernment for ID and OOD data, ultimately improving the OOD detection task.
The final optimization objective is defined as follows:
\begin{equation}
\label{eq:final_loss}
\mathcal{L} = \mathcal{L}_{\text{OMB}} + \alpha \cdot \mathcal{L}_{\text{OKP}},
\end{equation}
where $\alpha$ is a hyper-parameter, controlling the trade-off between learning new OOD semantics from pseudo-labeled data and suppressing the noise introduced by ID boundary samples.

\begin{table*}[!ht]
\centering
\renewcommand{\arraystretch}{1.15}
\setlength{\tabcolsep}{2.0pt}
\small 
\caption{Performance comparison on ImageNet-1k OOD detection benchmark. 
'$\checkmark$' indicates that the method uses additional resources such as large language models, external image data, or external text labels.
Lower FPR95 and higher AUROC are better. Best results are in \textbf{bold} and the second-best results are \underline{underlined}. }
\vspace{-6pt}
\label{tab:imagenet-1k}
\resizebox{0.95\textwidth}{!}{
\begin{tabular}{lcccccccccccc}
\toprule
\multirow{2}{*}{Method} & \multirow{2}{*}{Extra Resources} 
& \multicolumn{2}{c}{iNaturalist} 
& \multicolumn{2}{c}{SUN} 
& \multicolumn{2}{c}{Places} 
& \multicolumn{2}{c}{Texture} 
& \multicolumn{2}{c}{\textbf{Average}} \\
\cmidrule(lr){3-4} \cmidrule(lr){5-6} \cmidrule(lr){7-8} \cmidrule(lr){9-10} \cmidrule(lr){11-12}
& & FPR95$\downarrow$ & AUROC$\uparrow$ & FPR95$\downarrow$ & AUROC$\uparrow$ & FPR95$\downarrow$ & AUROC$\uparrow$ & FPR95$\downarrow$ & AUROC$\uparrow$ & FPR95$\downarrow$ & AUROC$\uparrow$ \\
\midrule
\multicolumn{11}{c}{\textit{Post-hoc methods}} \\
MCM~\citep{mcm}  &     & 30.92 & 94.61 & 37.59 & 92.57 & 44.71 & 89.77 & 57.85 & 86.11 & 42.77 & 90.76 \\
GL-MCM~\citep{gl-mcm} &   & 15.09 & 96.72 & 29.08 & 93.41 & 37.07 & 90.37 & 58.94 & 83.11 & 35.04 & 90.90 \\
CMA~\citep{cma} &  $\checkmark$      & 23.84 & 96.89 & 30.11 & 93.69 & 29.86 & 93.17 & 47.35 & 88.47 & 32.79 & 93.05 \\
CLIPN~\citep{wang2023clipn} &  $\checkmark$   & 19.17 & 96.17 & 26.43 & 94.02 & 32.26 & 92.62 & 41.23 & 90.12 & 30.21 & 93.19 \\
Neglabel~\citep{neglabel} &  $\checkmark$  & ~~2.00     & 99.47 & 20.95 & 95.47 & 36.48 & 91.56 & 45.00    & 90.02 & 26.10  & 94.13 \\  
NegRefine~\cite{negrefine} &  $\checkmark$   & ~~1.61     & 99.53 & 23.70 & 94.50 & 40.39 & 89.91 & \textbf{22.37}    & 94.39 & 22.01 & 94.58 \\  
CSP~\cite{csp} &  $\checkmark$   & ~~1.54     & 99.60 & 13.66 & 96.66 & 29.32 & 92.90 & \underline{25.52}    & 93.86 & \underline{17.51} & 95.76 \\  
\midrule
\multicolumn{11}{c}{\textit{Training-based methods}} \\
LoCoOp~\citep{locoop} &       & 16.05 & 96.86 & 23.44 & 95.07 & 32.87 & 91.98 & 42.28 & 90.19 & 28.66 & 93.52 \\
IDLike~\citep{idlike}  &      & 19.23 & 96.70 & 54.15 & 87.64 & 56.63 & 85.86 & 34.69 & 91.90 & 41.17 & 90.52 \\
Local-Prompt~\citep{zeng2025local} &  & ~~8.62 & 98.06 & 23.78 & 95.22 & 32.43 & 92.50 & 48.47 & 88.84 & 28.32 & 93.65 \\
APT~\citep{APT} &  $\checkmark$  & ~~9.70 & 97.79 & 20.12 & 95.52 & 28.54 & 92.84 & 45.78 & 88.43 & 26.03 & 93.64 \\
FA~\citep{fa}  &        & 13.37 & 96.80 & 28.83 & 93.12 & 30.30 & 92.54 & 30.50 & 92.66 & 25.75 & 93.78 \\
OSPCoOp~\citep{ospcoop} & $\checkmark$ & 15.25 & 97.13 & 18.26 & 96.74 & 25.74 & 94.01 & 41.26 & 91.13 & 25.13 & 94.75 \\
MoFE~\citep{mofe} &   & 5.19 & 97.28 & 22.10 & 95.17 & \underline{21.32} & \underline{94.69} & 31.47 & 92.15 & 20.02 & 94.89 \\
\midrule
\multicolumn{11}{c}{\textit{Test-time adaptation methods}} \\
OODD~\citep{oodd}  &      & 2.13  & 99.39 & 21.99 & 95.01 & 41.91 & 88.13 & 28.53 & 93.84 & 23.64 & 94.09 \\
AdaND~\citep{cao2025zsntta}  &       & 7.06  & 98.50  & 26.95 & 93.92 & 31.41 & 92.27 & 32.61 & 91.32    & 24.50 & 94.00 \\
AdaNeg~\citep{zhang2024adaneg}  &  $\checkmark$    & \underline{0.90}   & \underline{99.69} & \underline{11.57} & \underline{96.97} & 35.16 & 93.69 & 29.27 & \underline{94.34} & 19.22 & \underline{96.17} \\
\rowcolor{red!10}
TTL (Ours) &   & \textbf{0.42} & \textbf{99.87} & \textbf{7.18} & \textbf{98.45} & \textbf{15.86} & \textbf{96.22} & 26.39 & \textbf{94.60} & \textbf{12.46} & \textbf{97.29} \\
\bottomrule
\end{tabular}
}
\end{table*}

\noindent\textbf{OOD Textual Knowledge Bank.} Furthermore, a key challenge in test-time adaptation is that text prompts optimized within individual batches often capture only local semantic information, making them unstable to distribution shifts in subsequent batches. To overcome this limitation, we propose the OOD Textual Knowledge Bank (OKB), which accumulates discriminative textual features learned across test batches.
The OKB serves two critical purposes: (1) preserving valuable OOD textual knowledge that might be forgotten when adapting to new batches, and (2) providing broader semantic coverage through diverse distributions. 
%
By accumulating diverse OOD knowledge across batches, the OKB provides a more stable reference for calibration, leading to improved robustness under distribution shifts.
%
To maintain computational efficiency, the bank operates with a fixed capacity $K$. 
For each OOD prompt, its corresponding text feature vector $\mathbf{t}_i^{ood}$ is assigned a potential OOD score $S_{\text{in}}(\mathbf{t}_i^{ood})$, defined as the minimum distance to the text features of all ID prompts:
%
\begin{equation}
    \label{eq:s_in_score}
    S_{\text{in}}(\mathbf{t}_i^{ood}) = \min_{c} \left[-\cos(\mathbf{t}_c^{id}, \mathbf{t}_i^{ood}) \right].
\end{equation}
This score guides the bank's update strategy: when capacity is reached, the method retains only the $K$ highest-scoring features, ensuring the text features of the most discriminative OOD prompts are preserved.

\subsection{Calibration of the Prediction}
During inference, the OOD textual knowledge bank is used to calibrate the final prediction.
Specifically, for each test sample with image feature vector $\mathbf{z}$, the text features from the bank are used to compute the calibrated OOD detection score $S_{\text{cal}}(\mathbf{x})$ as follows,
\begin{align}
\quad S_{\text{cal}}(\mathbf{x}) = -\max_{j \in \{1, \ldots, K\}} \cos(\mathbf{z}, \mathbf{t}_j^{ood}) \,.
\label{eq:cal_score}
\end{align}
Then, the final OOD detection score can be defined as
\begin{align}
S_{\text{final}}(\mathbf{x}) = S_{\text{base}}(\mathbf{x}) + \beta \cdot S_{\text{cal}}(\mathbf{x}) \,,
\label{eq:final_score}
\end{align}
where $S_{\text{base}}(\mathbf{x})$ is the score from a base OOD detector, and fusion coefficient $\beta$ is a hyper-parameter, balancing semantic information and order of magnitude.
Benefiting from the $S_{\text{cal}}(\mathbf{x})$, the final scores for OOD data are effectively suppressed compared to those for ID data, thereby enhancing OOD detection performance.

\section{Experiments}

\begin{table*}[h]
\centering

\renewcommand{\arraystretch}{1.15}
\setlength{\tabcolsep}{5pt}
\small 

\caption{Performance comparison on the CIFAR-100 OOD benchmark. F and A denote FPR95 and AUROC, respectively.}
\vspace{-6pt}
\label{tab:cifar100}
\begin{tabular}{lcccccccccccccc}
\toprule
\multirow{2}{*}{Method} & \multicolumn{2}{c}{SVHN} & \multicolumn{2}{c}{LSUN-R} & \multicolumn{2}{c}{LSUN-C} & \multicolumn{2}{c}{iSUN} & \multicolumn{2}{c}{Texture} & \multicolumn{2}{c}{Places365} & \multicolumn{2}{c}{\textbf{Average}} \\
\cmidrule(lr){2-3} \cmidrule(lr){4-5} \cmidrule(lr){6-7} \cmidrule(lr){8-9} \cmidrule(lr){10-11} \cmidrule(lr){12-13} \cmidrule(lr){14-15}
& F$\downarrow$ & A$\uparrow$ 
& F$\downarrow$ & A$\uparrow$ 
& F$\downarrow$ & A$\uparrow$ 
& F$\downarrow$ & A$\uparrow$ 
& F$\downarrow$ & A$\uparrow$ 
& F$\downarrow$ & A$\uparrow$ 
& F$\downarrow$ & A$\uparrow$ \\
\midrule
MCM          & 67.72 & 89.49 & 67.49 & 86.51 & 47.56 & 91.97 & 65.90  & 86.48 & 92.09 & 72.98 & 97.79 & 60.97 & 73.09 & 81.40 \\
CSP & 99.40 & 70.95   & 66.43 & 89.81 & 78.82 & 85.97 & 66.39 & 89.57 & 26.04 & 92.59 & 88.99 & 67.22 & 71.01 & 82.68 \\
FA     & 15.58 & 97.33 & 48.02 & 89.75 & 33.11 & 93.18 & 51.29 & 89.37 & \underline{22.34} & \underline{95.47} & \underline{46.28} & \underline{89.49} & 36.11 & 92.43 \\
\midrule
OODD         & 60.60  & 90.52 & 75.20  & 84.61 & 44.72 & 92.08 & 75.98 & 83.85 & 90.85 & 73.12 & 98.09 & 59.09 & 74.24 & 80.55 \\
AdaND        & ~~\underline{1.17}  & \underline{99.67} & \underline{11.98} & \underline{96.76} & \underline{5.63}  & \underline{98.35} & \underline{18.71} & \underline{95.70}  & 22.52 & 92.56 & 65.71 & 72.00 & 20.95 & \underline{92.50} \\
AdaNeg       & 16.80  & 95.13 & 32.29 & 89.85 & 30.84 & 91.35 & 38.09 & 88.10  & 45.88 & 91.31 & 79.21 & 78.08 & 40.52 & 88.97 \\
\rowcolor{red!10}
TTL (Ours)  & \textbf{~~0.01} & \textbf{99.98} & \textbf{~~2.52} & \textbf{99.47} & \textbf{~~2.01} & \textbf{99.54} & \textbf{~~1.27} & \textbf{99.67} & \textbf{~~1.68} & \textbf{98.91} & \textbf{~~6.65} & \textbf{97.98} & \textbf{~~2.36} & \textbf{99.26} \\
\bottomrule
\end{tabular}
\end{table*}

\subsection{Experimental Setup}

\noindent\textbf{Datasets and Evaluation Protocol.} 
Following prior protocols~\citep{locoop, tagfog}, we evaluate our method on two standard benchmarks. 
For large-scale evaluation, ImageNet-1K~\citep{DBLP:conf/cvpr/DengDSLL009} was used as the ID dataset, with OOD test sets comprising iNaturalist~\citep{DBLP:conf/cvpr/HornASCSSAPB18}, SUN~\citep{DBLP:conf/cvpr/XiaoHEOT10}, Places~\citep{DBLP:journals/pami/ZhouLKO018}, and Texture~\citep{DBLP:conf/cvpr/CimpoiMKMV14}.
Considering broader comparison across different dataset scales and resolutions, we adopt CIFAR-100~\citep{cifar} as the ID dataset and six datasets serve as OOD datasets, including SVHN~\citep{SVHN}, LSUN-C~\citep{LSUN}, LSUN-R~\citep{LSUN}, iSUN~\citep{isun}, Texture~\citep{DBLP:conf/cvpr/CimpoiMKMV14}, and Places365~\citep{DBLP:journals/pami/ZhouLKO018}.
For evaluation, we adopt two standard metrics: (1) FPR95: false positive rate at 95\% recall; (2) AUROC: area under the receiver operating characteristic curve.


\begin{figure}[t] 


            \begin{center}
        \includegraphics[width=0.48\textwidth]{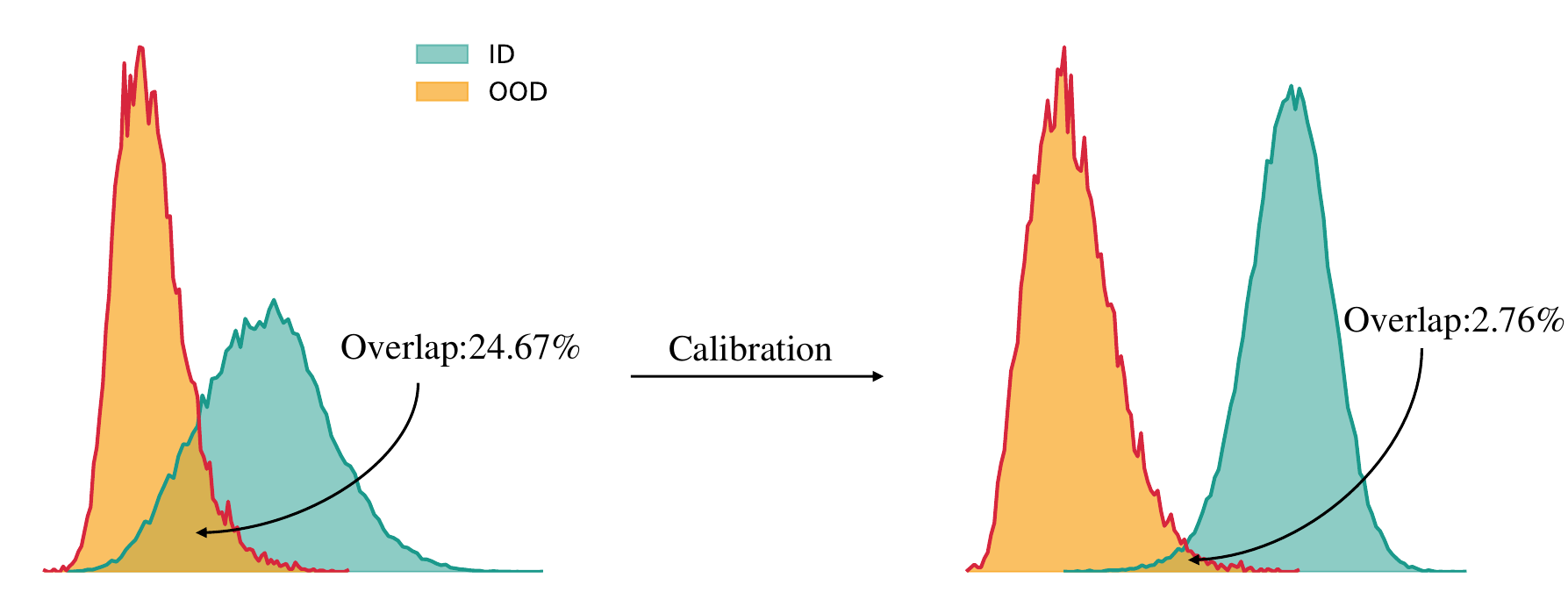}
        \end{center}
        \vspace{-10pt}
        \caption{Score density distributions for ID~(ImageNet) and OOD~(SUN) samples before and after calibration with our TTL, where the MCM scoring function is used as the base detector. }

        \label{fig:overview_calibration}

\end{figure}

\noindent\textbf{Implementation Details.} 
Following the setup used in prior work~\citep{zhang2024adaneg,owttt,oodd}, CLIP~\citep{clip} with ViT-B/16~\citep{DBLP:conf/iclr/DosovitskiyB0WZ21} is adopted as the backbone model, and MCM~\citep{mcm} is used as the base OOD detector.
The learnable OOD prompts are optimized using AdamW~\citep{DBLP:journals/corr/KingmaB14} with a learning rate of 0.005.
Four hyper-parameters for our TTL are set as : loss weight $\alpha=0.5$, , OKB capacity $K=2048$, batch size $B = 64$, and fusion coefficient $\beta$ is set to 0.006 for CIFAR-100, 0.0005 for ImageNet-1k. 
The $\beta$ is small, as the calibrated score and the base detector score lie on different order of magnitude.
More details are provided in the Appendix~\ref{app:setting}.
\noindent\textbf{Comparison Methods.}
Comparisons of TTL with competitive CLIP-based OOD detection methods are presented across three paradigms:
(1) post-hoc methods, including MCM~\cite{mcm}, GL-MCM~\cite{gl-mcm}, Neglabel~\cite{neglabel}, CMA~\cite{cma}, NegRefine~\cite{negrefine}, CSP~\cite{csp}, 
and CLIPN~\cite{wang2023clipn};
(2) training-based methods contains LoCoOp~\cite{locoop},  Local-Prompt~\cite{zeng2025local}, IDLike~\cite{idlike}, APT~\cite{APT},  OSPCoOp~\cite{ospcoop}, MoFE~\cite{mofe} and FA~\cite{fa}, which fine-tunes on ID training datasets;
and (3) test-time adaptation~(TTA) methods, including OODD~\cite{oodd}, AdaNeg~\cite{zhang2024adaneg} and AdaND~\cite{cao2025zsntta}. 


\subsection{Main Results}
\noindent\textbf{Effectiveness on ImageNet-1k Benchmark.} 
As shown in Table~\ref{tab:imagenet-1k}, TTL achieves the best OOD detection performance among all TTA methods.
Notably, unlike CSP and AdaNeg, which leverage OOD labels specifically tailored to the ImageNet-1K benchmark, TTL requires no predefined OOD knowledge.
Instead, it dynamically learns OOD knowledge during testing, resulting in superior performance (e.g., achieving an AUROC of 97.29\% and FPR95 of 12.46\% on average).
Moreover, without using any labeled ID data for training, TTL still outperforms the best training-based baseline MoFE, by 7.56\% in FPR95.
As shown in Figure~\ref{fig:overview_calibration}, by calibrating the extracted OOD textual knowledge, TTL achieves a clearly larger discrepancy between ID and OOD samples.
These results support that TTL's ability to learn OOD textual knowledge during testing leads to better OOD detection performance.


\noindent\textbf{Effectiveness on CIFAR-100 Benchmark.} 
The superior performance of our TTL is also confirmed on the CIFAR-100 benchmark.
As shown in Table~\ref{tab:cifar100}, TTL consistently outperforms all competing methods across various OOD datasets and achieves the best performance on average.
Specifically, among test-time adaptation methods, TTL demonstrates  strong performance on the challenging Places365 dataset, achieving an AUROC improvement of 19.9\% where other TTA methods fail to deliver effective OOD detection.


\subsection{Ablation Study and Sensitivity Analyses}

\noindent\textbf{Ablation Study of $\mathcal{L}_{\text{OMB}}$, $\mathcal{L}_{\text{OKP}}$ and OKB.} 
Ablation studies are conducted on both ImageNet-1k and  CIFAR-100 OOD benchmarks to validate the key components of TTL: $\mathcal{L}_{\text{OMB}}$, $\mathcal{L}_{\text{OKP}}$, and OKB. 
As shown in Table~\ref{tab:ablation_ablation}, the OOD detection performance is gradually improved with more components are included, which confirms the effectiveness of each proposed component.
Notably, despite the strong performance achieved by the pseudo–label–guided objective $\mathcal{L}_{\rm OMB}$ and texutal bank OKB, integrating the OOD knowledge purification objective $\mathcal{L}_{\rm OKP}$ brings additional improvements (e.g.,  an average AUROC increase of +1.03\% across the two benchmarks), indicating that alleviating the effect of noisy pseudo labels is crucial for robust OOD detection.

\begin{table}[htbp]
        \centering
        \caption{Ablation study on the key components $\mathcal{L}_{\rm OMB}$, $\mathcal{L}_{\rm OKP}$ and OKB across two standard benchmarks.} 
        \setlength{\tabcolsep}{2.2pt}
        \small 
        \vspace{-6pt}
        \label{tab:ablation_ablation}
        \begin{tabular}{ccc|cccc}
        \toprule
        \multirow{2}{*}{ $\mathcal{L}_{\text{OMB}}$} & \multirow{2}{*}{ $\mathcal{L}_{\text{OKP}}$} & \multirow{2}{*}{OKB} & \multicolumn{2}{c}{ImageNet-1k} & \multicolumn{2}{c}{CIFAR-100}  \\
        \cmidrule(lr){4-5} \cmidrule(lr){6-7}
        & & & FPR95$\downarrow$ & AUROC$\uparrow$ & FPR95$\downarrow$ & AUROC$\uparrow$  \\
        \midrule
        \rowcolor{gray!20}
           \XSolidBrush         &   \XSolidBrush   & \XSolidBrush     & 42.77  & 90.76 & 73.09 & 81.40 \\
        $\checkmark$ &   \XSolidBrush   & \XSolidBrush     & 30.56 & 92.54 & 14.40 & 96.25\\
        $\checkmark$ & $\checkmark$ &   \XSolidBrush    & 24.59 & 93.95 & ~~4.14 & 98.71\\
        $\checkmark$ & \XSolidBrush     &   $\checkmark$  & 18.40 & 95.63 & ~~5.23 & 98.86\\
        $\checkmark$ & $\checkmark$ & $\checkmark$ & \textbf{12.46} & \textbf{97.29} & \textbf{~~2.36} & \textbf{99.26}\\
        \bottomrule
        \end{tabular}
        
         

\end{table}

\begin{figure}[h] 

        \begin{center}
        \includegraphics[width=0.47\textwidth]{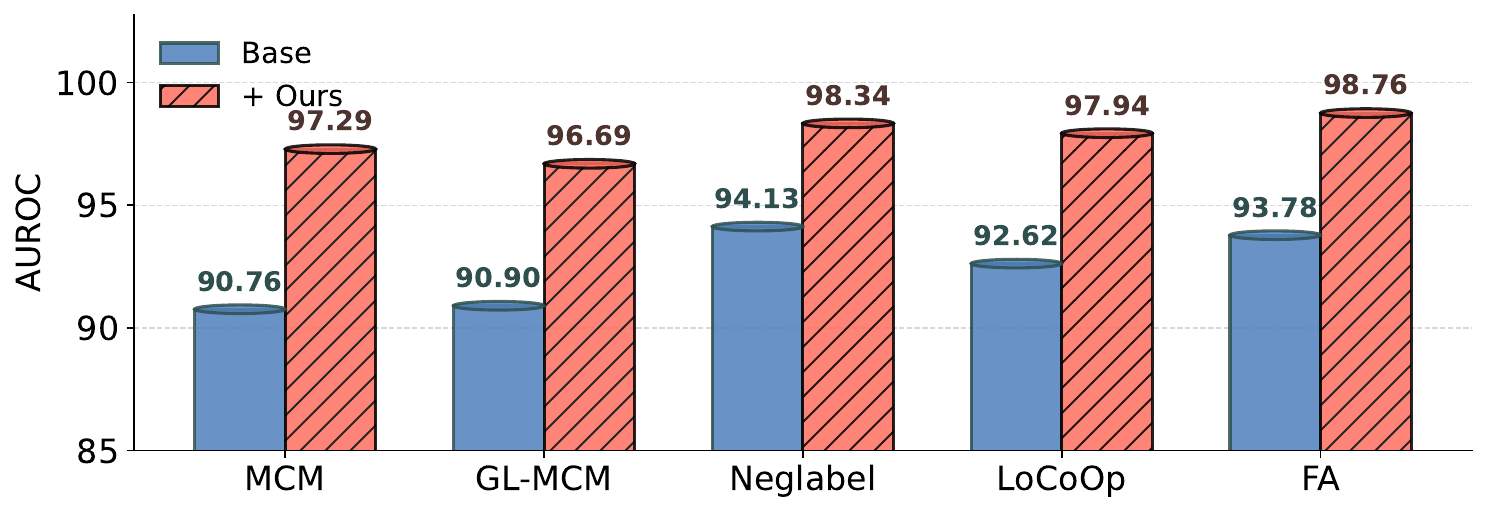}
        \end{center}
        \vspace{-12pt}
        \caption{Performance when integrated with different detectors.}
        \label{fig:overlay_with_other_method}

    
\end{figure}

\noindent\textbf{Sensitivity of Basic OOD Detector.}
As Figure~\ref{fig:overlay_with_other_method} shows, TTL improves performance across all base OOD detectors on the ImageNet-1k benchmark.
The gains are particularly pronounced when our TTL is combined with FA, achieving 5.88\% FPR95 and 98.76\% AUROC.
The study shows the insensitivity to the selection of base OOD detetcor for TTL. 

\noindent\textbf{Update Strategies of OKB.}
Additional three OKB update strategies are evaluated on both ImageNet-1k and CIFAR-100 benchmarks: 
(1) Random replacement when OKB is full~\citep{rand} (RAND); 
(2) First-In-First-Out (FIFO)~\citep{fifo}; 
(3) Storing all (SA)~\citep{sa}, assuming unlimited retention. 
The proposed update strategy achieves the best performance on both benchmarks.
The results indicate that updating the memory with the learned OOD prompts that has the lowest similarity to the ID textual features enables more accurate collection of OOD knowledge during testing across different ID and OOD datasets.

\begin{table}[t] 
 
        \centering
\setlength{\tabcolsep}{4.0pt}
\small 
\captionof{table}{Comparison of OKB update strategies.}
\vspace{-6pt}
\label{tab:cache_update_stra}
\begin{tabular}{lcccc}
\toprule
\multirow{2}{*}{Strategy} & \multicolumn{2}{c}{ImageNet-1k} & \multicolumn{2}{c}{CIFAR-100} \\
\cmidrule(lr){2-3} \cmidrule(lr){4-5}
 & FPR95$\downarrow$ & AUROC$\uparrow$ & FPR95$\downarrow$ & AUROC$\uparrow$  \\
\midrule
 RAND & 27.29 & 93.07 & 29.06 & 87.07 \\
 FIFO  & 14.69  & 96.40 & ~~8.15 & 98.78 \\
 SA & 23.19  & 94.27 & 27.33 & 88.04 \\
 \rowcolor{red!10}
 Ours & \textbf{12.46}  & \textbf{97.29} & \textbf{~~2.36} & \textbf{99.26}\\
\bottomrule
\end{tabular}
\end{table}

\begin{figure*}[htbp] 

        
         
        \begin{center}
        \includegraphics[width=0.98\textwidth]{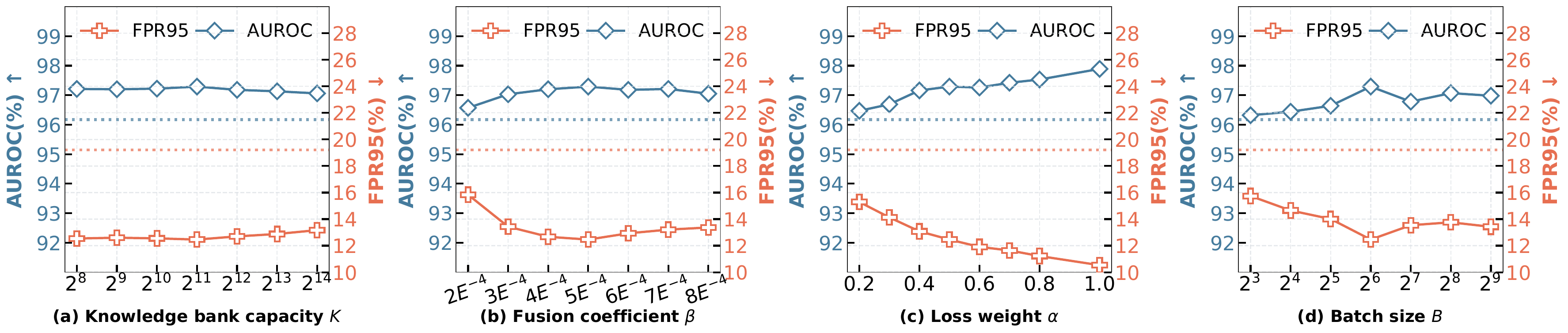}
        \end{center}
        \vspace{-13pt}
        \caption{Hyper-parameters sensitivity studies on ImageNet-1k benchmark. Dashed lines represent the performance of AdaNeg.}
        \label{fig:sensitivity}

\end{figure*}






\noindent\textbf{Initialization of OOD Prompt.} 
Table \ref{tab:ood_prompt_initialization} compares several OOD-prompt initialization strategies on both ImageNet-1k  and CIFAR-100 benchmarks.
While the frozen parts of OOD prompt are initialized by ID class names, this initialization strategy promotes the exploration of OOD knowledge that lies close to ID semantics, thereby enhancing the model’s OOD detection capability.
Moreover, the initialization strategy based on fully hand-crafted prompts (i.e., ``a photo of a [\textit{classname}].") yields the best OOD detection performance, possibly because such manually designed prompts facilitate faster test-time adaptation of the model.
Overall, the results show that employing manual initialization for OOD prompts enables effective adaption during testing and mining the critical OOD knowledge.

\begin{table}[h] 
        \centering
        \setlength{\tabcolsep}{2.2pt}
        \small 
        \captionof{table}{Comparison of OOD prompt learning strategies. For Prefix, `$\checkmark$' indicates using ``a photo of a" initialization, and `\XSolidBrush' indicates random initialization. For Classname, `$\checkmark$' indicates initialization using ID labels, and `\XSolidBrush' indicates random initialization. }
        \vspace{-6pt}
        \label{tab:ood_prompt_initialization}
        \begin{tabular}{cc|cccc}
        \toprule
        \multirow{2}{*}{Prefix} & \multirow{2}{*}{Classname} & \multicolumn{2}{c}{ImageNet-1k} & \multicolumn{2}{c}{CIFAR-100} \\
        \cmidrule(lr){3-4} \cmidrule(lr){5-6}
         & & FPR95$\downarrow$ & AUROC$\uparrow$ & FPR95$\downarrow$ & AUROC$\uparrow$  \\
        \midrule
         \XSolidBrush & \XSolidBrush &  15.02  & 96.50 & 45.23 &  88.39\\
         $\checkmark$ & \XSolidBrush & 13.48  & 96.88  & 45.29 & 88.45\\
         \XSolidBrush & $\checkmark$ &  14.21  & 96.69  & ~~4.92 &  99.03\\
         \rowcolor{red!10}
         $\checkmark$ & $\checkmark$ & \textbf{12.46}  & \textbf{97.29}  & ~~\textbf{2.36}  & \textbf{99.26}\\
         
        \bottomrule
        \end{tabular}
\end{table}

\noindent\textbf{Sensitivity Study on Hyper-parameters.} 
Sensitivity studies are conducted to assess the impact of the hyperparameters in our TTL, where the knowledge bank capacity $K \in [2^8, 2^{14}]$, fusion coefficient $\beta \in [0.0002, 0.0008]$,  loss weight $\alpha \in [0.2, 1.0]$, and batch size $B \in [2^3, 2^{9}]$. 
As shown in Figure~\ref{fig:sensitivity}, our method is largely insensitive to hyperparameter choices and consistently surpasses the strongest baseline.
Moreover, under diverse test-time environments induced by varying batch sizes, our TTL maintains superior performance, demonstrating strong robustness and adaptability.

\subsection{Further Discussion}

\begin{figure}[htbp] 
             \begin{center}
        \includegraphics[width=0.4\textwidth]{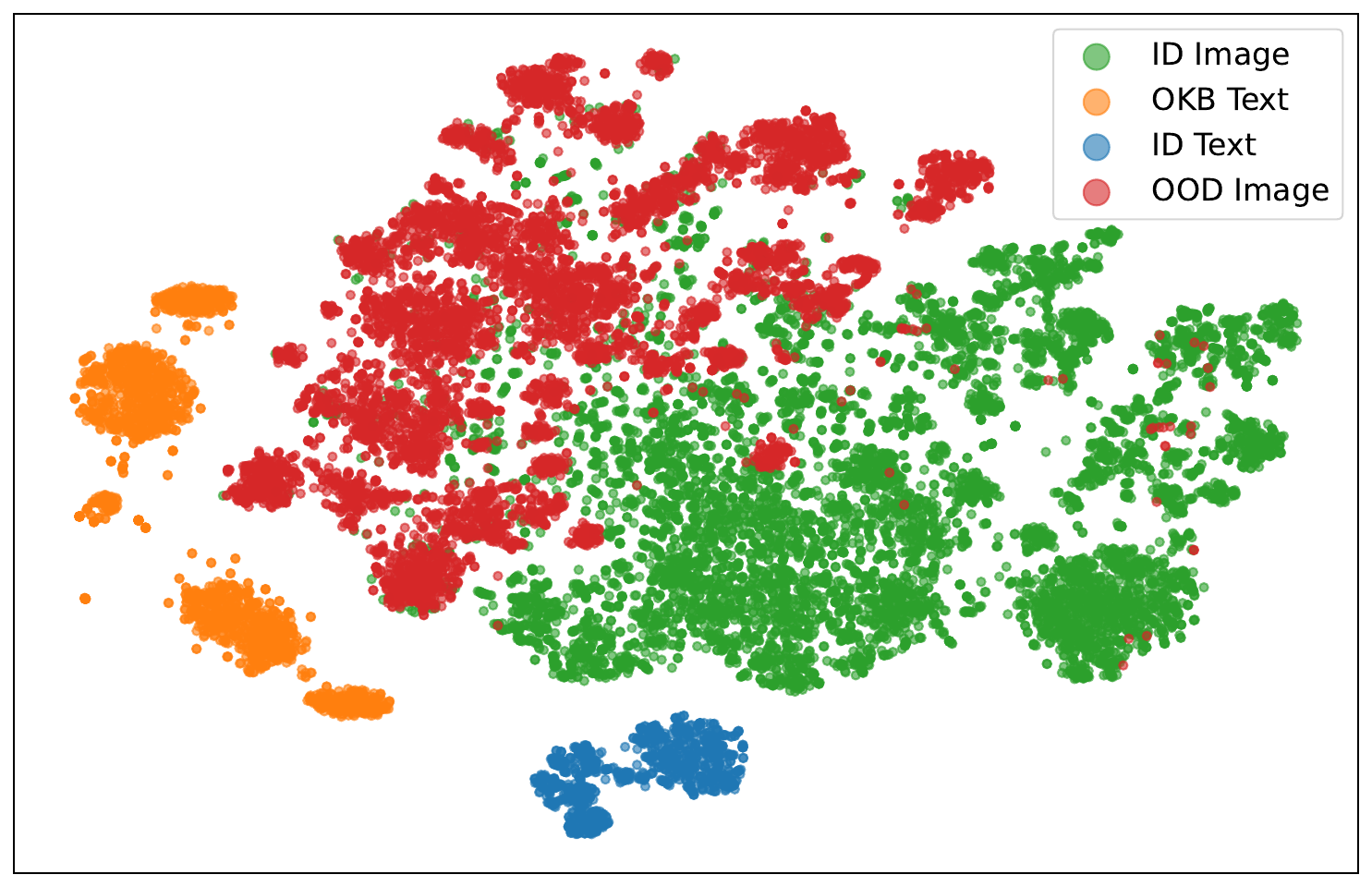}
        \end{center}
        \vspace{-10pt}
        \caption{t-SNE visualization alongside of textual embeddings in OKB, where ImageNet-1k serves as ID dataset, and SUN as OOD dataset respectively.}
        \label{fig:tsne}

\end{figure}

\noindent\textbf{Visualization of OKB.} 
Figure \ref{fig:tsne} presents a t-SNE visualization for both ID samples and OOD samples.
We observe that the learned OOD prompts tend to cluster closer to real OOD samples in the CLIP embedding space, confirming that our TTL effectively capture OOD semantics by OOD prompts as intended.

\noindent\textbf{Effect of Textual and Visual Knowledge Banks.}
While our TTL outperforms visual-only bank methods~\cite{oodd,zhang2024adaneg} by
learning OOD prompts on the textual encoder side, we also analyze the effect of integrating it with a visual-only bank.
Specifically, the variant \textbf{TTL-V} is introduced by combining the TTL score with the score derived from a visual bank, following the design of OODD~\cite{oodd}.
As shown in Table~\ref{tab:bank_comparison}, TTL-V brings further performance gains by introducing a visual bank, but at the expense of increased storage.
Overall, while TTL performs test-time adaptation on the textual encoder side alone, TTL is also compatible with previous methods that adapt based on visual features.


\begin{table}[h]
\centering
\setlength{\tabcolsep}{2.5pt}
\small 
\caption{Comparison between visual-only, textual-only, and both knowledge banks on the ImageNet-1k benchmark. }
\vspace{-6pt}
\begin{tabular}{l|cc|ccc}
\toprule
Method & Visual & Text & FPR95 $\downarrow$ & AUROC $\uparrow$  & Storage $\downarrow$\\
\midrule
OODD & $\checkmark$ &  \XSolidBrush & 23.64 & 94.09 & 4.00MB\\
TTL & \XSolidBrush & $\checkmark$  & 12.46 & 97.29 & 4.00MB\\
TTL-V & $\checkmark$ & $\checkmark$ & ~~9.19 & 98.04 & 8.00MB\\
\bottomrule
\end{tabular}
\label{tab:bank_comparison}
\end{table}

\noindent\textbf{Storage and Time Analysis.} 
To evaluate our TTL's additional computational cost for adaptation during testing, we compared its runtime and additional storage usage with the other methods.
As shown in Table~\ref{tab:time_complexity}, although training-based and post-hoc methods offer faster inference, their OOD detection is typically inferior by the TTA methods due to the lack of adaptation to test samples.
Our TTL incurs marginally higher inference cost than prior TTA methods because they typically perform adaptation based on the encoder outputs, whereas our approach updates OOD prompts at the input side.
To mitigate the computational cost, an early-stopping strategy can be employed for OOD prompt updates during testing. Specifically, when the updates stopped after 1,280 test samples, the model attains the FPR95 of 14.21\%, remaining superior to the best TTA baseline AdaNeg by 5.01\%.
In summary, our method delivers substantial performance gains at a modest computational cost. 
Furthermore, the integration of cost-saving strategies, such as early stopping, can be employed to further improve efficiency during test-time optimization.

\begin{table}[htbp]
  \centering
  \caption{Comparison of the storage and runtime usage. ``Storage" indicates the extra memory occupation, ``Training" denotes the training time,  and ``Testing" denotes per-image inference time. Experiments are conducted with a single NVIDIA 3090 GPU.}
  \vspace{-6pt}
  \resizebox{0.46\textwidth}{!}{
  \label{tab:time_complexity}
  \begin{tabular}{l|cccccc}
    \toprule
    \textbf{Methods}  & \textbf{Storage} & \textbf{Training} & \textbf{Testing} & \textbf{FPR95 $\downarrow$} \\
    \midrule
    MCM       & --   & --   &  ~8.18ms   & 42.77 \\
    FA         & --     & 2h    & ~8.20ms   & 25.75 \\
    \midrule
    OODD            & ~~~~4.00MB  & --   & ~8.23ms   & 23.64 \\
    AdaND          & ~~~~0.25MB   & --   & 10.72ms     & 24.50 \\
    AdaNeg          & 214.75MB & --   & 10.27ms  & 19.22 \\
    \rowcolor{red!10}
    \textbf{TTL~(Ours)}  & ~~~~4.00MB & -- & 11.40ms & 12.46\\
    \rowcolor{red!10}
    \textbf{+~early stopping}  & ~~~~4.00MB & -- & ~8.36ms & 14.21\\
    \bottomrule 
  \end{tabular}
  }
\end{table}
\section{Conclusion}

In this work, we propose TTL, a new framework that adapts vision-language models for OOD detection by dynamically learning OOD textual semantics from unlabeled test streams. Unlike existing methods that rely on fixed labels, TTL incrementally constructs OOD textual knowledge that better reflects evolving test distributions.
To mitigate unavoidable noise on adaptation, we propose an OOD knowledge purification strategy, enabling the learned knowledge to acquire clearer OOD semantics. Moreover, we developed an OOD Textual Knowledge Bank to stabilize score calibration across batches and further enhance detection performance.
Extensive experiments demonstrate that TTL consistently achieves state-of-the-art OOD detection results, validating the value of textual adaptation in test-time scenarios for robust test-time OOD detection.


\section*{Acknowledgements}
This work is supported in part by the National Natural Science Foundation of China (grant No. 62571559), the Major Key Project of PCL (grant No. PCL2025AS209), and Guangdong Excellent Youth Team Program (grant No. 2023B1515040025).

{
    \small
    \bibliographystyle{ieeenat_fullname}
    \bibliography{main}
}


\clearpage
\setcounter{page}{1}
\maketitlesupplementary
\renewcommand\thesection{\Alph{section}}
\setcounter{section}{0}

\section{Basic statement}

\subsection{The Use of Large Language Models} 
Throughout the entire work, we use ChatGPT for language polishing and code assistance.

\subsection{Reproducibility Statement} we will make the complete implementation code, including model training configurations and evaluation scripts, publicly available on GitHub upon the acceptance of this paper. Additionally, we will provide detailed documentation specifying dependencies and step-by-step instructions to replicate our experiments. This ensures that researchers can easily verify our findings and build upon our method for future work.


\subsection{Detailed experiment setting}
\label{app:setting}

\noindent\textbf{Experimental Details} The proposed TTOD method was implemented using Python 3.9 and PyTorch 2.3.0, with all experiments conducted on a single NVIDIA GeForce RTX 3090 GPU. Following prior work~\citep{zeng2025local,tpt, owttt,mcm, oodd}, we adopted ViT-B/16~\citep{DBLP:conf/iclr/DosovitskiyB0WZ21} as the backbone model. The OOD prompt was optimized via AdamW~\citep{DBLP:journals/corr/KingmaB14} with a learning rate of 0.005 and batch size of 64. We set $N_q=512$ and $\tau=1$, and used MCM~\citep{mcm} as the base OOD detector across all experiments.  

TTL includes four hyperparameters: (1) 
Increasing the loss weight generally improves performance. However, very large weights over-emphasize pseudo-labeled OOD samples. From a practical deployment perspective, assuming an extremely unreliable base detector is unrealistic. Therefore, we select a moderate weight (i.e., $\alpha = 0.5$), which emphasizes OOD knowledge learning without overly trusting noisy pseudo-labels—even though this is not the weight that yields the highest peak performance; (2) OKB capacity: TTL is highly robust to the choice of OKB size. To ensure fair comparison with OODD—which maintains a visual memory bank—we adopt the same capacity, setting $K=2048$; (3) A larger batch size provides more comprehensive pseudo-label signals within each update, enabling more stable adaptation and generally improving detection performance. It also accelerates inference because more samples can be processed in parallel. However, when the batch size becomes excessively large, the number of adaptation steps per test stream decreases dramatically, leading to slower learning of new OOD semantics. This ultimately harms performance in streaming evaluation. To balance adaptation quality, learning efficiency, and inference throughput, we adopt a moderate batch size of $B=64$.; and (4) The fusion coefficient $\beta$ differs across datasets mainly because the base detector scores have different magnitudes. On CIFAR-100 and ImageNet-1k, the MCM scores vary by roughly one order of magnitude, and accordingly, the chosen fusion coefficients also differ by about one order of magnitude. Specifically, $\beta$ is set to 0.006 for CIFAR-100 and 0.0005 for ImageNet-1k.

\begin{algorithm}[htbp]
\caption{Test-time Textual OOD Learning}
\label{alg:ood_detection}
\begin{algorithmic}[1] 
\REQUIRE  
    test data stream $\{x_i\}_{i=1}^T$, text encoder $g(\cdot)$, image encoder $f(\cdot)$, text prompt for ID $u_{c}^{id}$, learnable text prompt for OOD $u_{c}^{ood}$, batch size $B$, the priority queue $OKB$ with capacity $K$


\STATE Initialize OOD prompt: $u_c^{ood}=u_c^{id}$
\STATE Compute and obtain ID text embeddings: $t_c^{id}=g(u_c^{id})$

\FOR{each data sample $x_i \in \mathcal{D}$}   
    \STATE Calculate base OOD score for $x_i$ using the base detector: $s_{\text{base}}(x_i)$
    \STATE Compute adaptive threshold: $\lambda$ by Equal~\ref{eq:adaptive_threshold}
    \STATE Assign pseudo-label to $x_i$: $\hat{y}$
    \STATE Obtain current OOD text embeddings: $t_c^{ood}=g(u_c^{ood})$
    \STATE Compute OOD probability $p$ by Equal~\ref{eq:ood_prob}
    \STATE Update queue $Q$:  $\mathcal{Q} \gets \mathcal{Q} \cup \{\hat{y}, p\}$
    \IF{$len(Q) = B$}  
        \STATE train $u_{c}^{ood}$:
        \STATE Calculate loss $\mathcal{L}_{\text{OMB}}$ loss by Equal~\ref{eq:cbce_loss}, input data: $Q$
        \STATE Compute grouping threshold $\theta$ by Equal~\ref{eq:adaptive_threshold}
        \STATE Calculate loss  $\mathcal{L}_{\text{OKP}}$ by Equal~\ref{eq:ditance_loss}, input data: $Q,\theta$
        \STATE Update $u_{c}^{ood}$ using  $\mathcal{L}_{\text{OMB}}$ and $\mathcal{L}_{\text{OKP}}$
        \STATE Obtain updated OOD prompt embeddings: $t_c^{ood}=g(u_c^{ood})$
        \STATE Score the updated OOD prompt embeddings by Equal~\ref{eq:s_in_score}
        \STATE Store each OOD prompt embedding into the priority queue $OKB$
        \STATE $Q \gets \emptyset$
    \ENDIF

    \STATE Perform inference correction and prediction:
    \STATE Compute final OOD score $S_{final}(x_i)$ by Equal~\ref{eq:final_score}
\ENDFOR  

\RETURN $S_{final}(x_i)$ for all samples in $\mathcal{D}$
\end{algorithmic}
\end{algorithm}

\noindent\textbf{Calibration Strategies.}  Several calibration strategies were adopted, and the specific formulas are shown in Table~\ref{tab:ood_knowledge_stra_all}.


\noindent\textbf{More Details about Adaptive Threshold.} OWTTT~\citep{DBLP:conf/iccv/Li0SJ23} searches for the optimal parameter $\lambda$ using a fixed step size of 0.01 between 0 and 1, which is actually unsuitable for different OOD scores (e.g., using ImageNet as ID data and the SUN dataset as OOD data, a step size of 0.01 could span the entire range of MCM Scores across all samples). Here, we propose using the minimum score encountered during testing as the lower bound for the search and the maximum score as the upper bound. Within this range, we uniformly divide the interval into segments matching OWTTT's approach to search for the optimal parameter $\lambda$.


\section{Algorithms}
This section provides a detailed breakdown of Algorithm \ref{alg:ood_detection} (Test-time Textual OOD Learning), complementing the core description in the main text. The algorithm aims to dynamically adapt OOD text prompts during testing, leveraging textual information to enhance OOD detection performance without relying on external pre-defined OOD categories.

\section{Full Results of Ablation Studies}

\noindent\textbf{Sensitivity of Basic OOD Detector.} Please refer to Table~\ref{tab:overlay_with_other_method}. We can see that we have achieved a positive improvement in performance for different basic detectors. And it's for all datasets, not just the average results. The study shows the insensitivity to the selection of base OOD detetcor for TTL.

\section{Additional Results}

\definecolor{softgreen}{RGB}{34,170,85}
\definecolor{softred}{RGB}{178,34,34}     

\begin{table*}[htbp]
\centering
\renewcommand{\arraystretch}{1.15}
\setlength{\tabcolsep}{3pt}
\small 
\caption{Complementarity to other OOD detectors with the ID dataset of ImageNet-1k. {\color{softgreen}Green} indicates an improvement, while {\color{softred}red} indicates the opposite.}
\vspace{-8pt}
\label{tab:overlay_with_other_method}
\resizebox{1.0\textwidth}{!}{
\begin{tabular}{lcccccccccccc}
\toprule
\multirow{2}{*}{Method}  
& \multicolumn{2}{c}{iNaturalist} 
& \multicolumn{2}{c}{SUN} 
& \multicolumn{2}{c}{Places} 
& \multicolumn{2}{c}{Texture} 
& \multicolumn{2}{c}{\textbf{Average}} \\
\cmidrule(lr){2-3} \cmidrule(lr){4-5} \cmidrule(lr){6-7} \cmidrule(lr){8-9} \cmidrule(lr){10-11}
& FPR95$\downarrow$ & AUROC$\uparrow$ 
& FPR95$\downarrow$ & AUROC$\uparrow$ 
& FPR95$\downarrow$ & AUROC$\uparrow$ 
& FPR95$\downarrow$ & AUROC$\uparrow$ 
& FPR95$\downarrow$ & AUROC$\uparrow$ \\
\midrule
MCM     
& 30.92 & 94.61 
& 37.59 & 92.57 
& 44.71 & 89.77 
& 57.85 & 86.11 
& 42.77 & 90.76 \\

+ \textbf{Ours} 
& \textbf{~~0.42} & \textbf{99.87} 
& \textbf{~~7.18} & \textbf{98.45}  
& \textbf{15.86}  & \textbf{96.22} 
& \textbf{26.39}  & \textbf{94.60} 
& \textbf{12.46} & \textbf{97.29} \\

Improve 
& {\color{softgreen} -30.50} & {\color{softgreen} +5.26} 
& {\color{softgreen} -30.41} & {\color{softgreen} +5.88}  
& {\color{softgreen} -28.85}  & {\color{softgreen} +6.45} 
& {\color{softgreen} -31.46}  & {\color{softgreen} +8.49} 
& {\color{softgreen} -30.31} & {\color{softgreen} +6.53} \\
\midrule

GL-MCM     
& 15.09 & 96.72 
& 29.08 & 93.41 
& 37.07 & 90.37 
& 58.94 & 83.11 
& 35.04 & 90.90 \\

+ \textbf{Ours} 
& \textbf{~~0.42} & \textbf{99.88} 
& \textbf{~~7.71} & \textbf{98.37}  
& \textbf{16.14}  & \textbf{96.04} 
& \textbf{33.67}  & \textbf{92.10} 
& \textbf{14.49} & \textbf{96.6} \\

Improve 
& {\color{softgreen} -14.67} & {\color{softgreen} +3.16} 
& {\color{softgreen} -21.37} & {\color{softgreen} +5.67}  
& {\color{softgreen} -20.93}  & {\color{softgreen} +5.67} 
& {\color{softgreen} -25.27}  & {\color{softgreen} +8.99} 
& {\color{softgreen} -20.55} & {\color{softgreen} +5.70} \\

\midrule
Neglabel     
& ~~2.00& 99.47 
& 20.95 & 95.47 
& 36.48 & 91.56 
& 45.00 & 90.02 
& 26.10  & 94.13 \\  

+ \textbf{Ours} 
& \textbf{~~0.44} & \textbf{99.86} 
& \textbf{6.49} & \textbf{98.68}  & \textbf{15.26}  & \textbf{96.81} & \textbf{10.74}  & \textbf{98.00} & \textbf{~~8.23} & \textbf{98.34} \\
Improve 
& {\color{softgreen} -1.56} & {\color{softgreen} +0.39} 
& {\color{softgreen} -14.46} & {\color{softgreen} +3.21}  
& {\color{softgreen} -21.22}  & {\color{softgreen} +5.25} 
& {\color{softgreen} -34.26}  & {\color{softgreen} +7.98} 
& {\color{softgreen} -17.87} & {\color{softgreen} +4.21} \\
\midrule
LoCoOp     
& 23.24 & 95.27 
& 31.56 & 93.76 
& 38.55 & 91.19 
& 43.43 & 90.28 
& 34.19 & 92.62 \\

+ \textbf{Ours} 
& \textbf{~~0.22} & \textbf{99.93} 
& \textbf{~~4.92}  & \textbf{98.88}  
& \textbf{13.91} & \textbf{96.51}  
& \textbf{16.12} & \textbf{96.44} 
& \textbf{~~8.79} & \textbf{97.94} \\
Improve 
& {\color{softgreen} -23.02} & {\color{softgreen} +4.66} 
& {\color{softgreen} -26.64} & {\color{softgreen} +5.32}  
& {\color{softgreen} -24.64}  & {\color{softgreen} +5.32} 
& {\color{softgreen} -27.31}  & {\color{softgreen} +6.16} 
& {\color{softgreen} -25.40} & {\color{softgreen} +5.32} \\
\midrule

FA              
& 13.37 & 96.80 & 28.83 & 93.12 & 30.30 & 92.54 & 30.50 & 92.66 & 25.75 & 93.78 \\
+ \textbf{Ours} & \textbf{0.28} & \textbf{99.90} & \textbf{5.87} & \textbf{98.68}  & \textbf{10.20}  & \textbf{97.81} & \textbf{7.15}  & \textbf{98.66} & \textbf{5.88} & \textbf{98.76} \\
Improve & {\color{softgreen} -13.09} & {\color{softgreen} +3.10} & {\color{softgreen} -22.96} & {\color{softgreen} +5.56}  & {\color{softgreen} -20.10}  & {\color{softgreen} +5.27} & {\color{softgreen} -23.35}  & {\color{softgreen} +6.00} & {\color{softgreen} -19.87} & {\color{softgreen} +4.98} \\
\bottomrule
\end{tabular}
}
\end{table*}
\begin{figure}[t]
\centering
\centering
\captionof{table}{Comparison of different calibration strategies.}
\resizebox{0.5\textwidth}{!}{
\label{tab:ood_knowledge_stra_all}
\begin{tabular}{lccc}
\toprule
Function & Formula  & FPR95$\downarrow$ & AUROC$\uparrow$  \\
\midrule
MaxSim & $-  \max_{j \in \{1, \ldots, K\}} \cos(\mathbf{z}_x, \mathbf{t}_j^{\text{ood}})$ & 20.52 & 95.57 \\
ExpSum & $-\sum_{j=1}^{K}\exp\left(\cos(\mathbf{z}, \mathbf{t}_j^{\text{ood}}) / \tau\right)$ & 19.61 & 95.44  \\
IDR  & $\frac{\sum_{j=1}^{N}\exp\left(\cos(\mathbf{z}, \mathbf{t}_j^{\text{id}}) / \tau\right)}
{\sum_{j=1}^{N} \exp\left(\cos(\mathbf{z}, \mathbf{t}_j^{\text{id}}) / \tau\right) + \sum_{j=1}^{K} \exp\left(\cos(\mathbf{z}, \mathbf{t}_j^{\text{ood}}) / \tau\right)}$  & 22.85 & 94.74 \\
\rowcolor{red!10}
Ours & $S_{\text{base}}(\mathbf{x}) - \beta \cdot S_{\text{cal}}(\mathbf{x})$    & \textbf{12.46} & \textbf{97.29} \\
\bottomrule
\end{tabular}
}

\end{figure}

\begin{table*}[htbp]
\centering
\renewcommand{\arraystretch}{1.15}
\setlength{\tabcolsep}{3pt}
\small 
\caption{Performance comparison on the ImageNet-1k benchmark with ResNet50 backbone.}
\vspace{-5pt}
\label{tab:different_arch}
\resizebox{\textwidth}{!}{
\begin{tabular}{lcccccccccccc}
\toprule
\multirow{2}{*}{Method}  
& \multicolumn{2}{c}{iNaturalist} 
& \multicolumn{2}{c}{SUN} 
& \multicolumn{2}{c}{Places} 
& \multicolumn{2}{c}{Texture} 
& \multicolumn{2}{c}{\textbf{Average}} \\
\cmidrule(lr){2-3} \cmidrule(lr){4-5} \cmidrule(lr){6-7} \cmidrule(lr){8-9} \cmidrule(lr){10-11}
& FPR95$\downarrow$ & AUROC$\uparrow$ & FPR95$\downarrow$ & AUROC$\uparrow$ & FPR95$\downarrow$ & AUROC$\uparrow$ & FPR95$\downarrow$ & AUROC$\uparrow$ & FPR95$\downarrow$ & AUROC$\uparrow$ \\
\midrule
MCM      & 32.92 & 93.73 & 47.73 & 90.43  & 60.67  & 85.71 & 61.65  & 85.24 & 50.74 & 88.78 \\
Neglabel & 2.60 & 99.29 & 22.62 & 95.05  & 47.71  & 90.0 & 42.85  & 89.80 & 28.95 & 93.53 \\
FA & 68.97 & 82.02 & 55.84 & 86.93  & 62.58  & 82.68 & 34.96  & 91.74 & 55.59 &  85.84 \\
\midrule
OODD       & 3.72 & 99.05 & 30.80 &  93.66 & 60.06  & 83.76 & 48.58  & 88.55 & 35.79 & 91.25 \\
AdaND      & 14.92 & 96.96 & 40.47 & 91.19  & 44.54  & 88.26 & 34.22  & 90.98 & 33.53 & 91.84 \\
AdaNeg     & \underline{1.07} & \underline{99.63} & \underline{13.22} & \underline{96.78}  & \underline{35.11}  & \underline{93.44} & \textbf{25.61}  & \textbf{94.67} & \underline{18.75} & \underline{96.13} \\
\rowcolor{red!10}
TTL(ours)     & \textbf{0.73}  & \textbf{99.75} & \textbf{9.39} & \textbf{98.16} & \textbf{18.49} & \textbf{95.25} & \underline{26.88} & \underline{94.21} & \textbf{13.87} & \textbf{96.84} \\
\bottomrule
\end{tabular}
}
\end{table*}

\begin{figure*}[ht]
    \centering
    \includegraphics[width=1.0\textwidth]{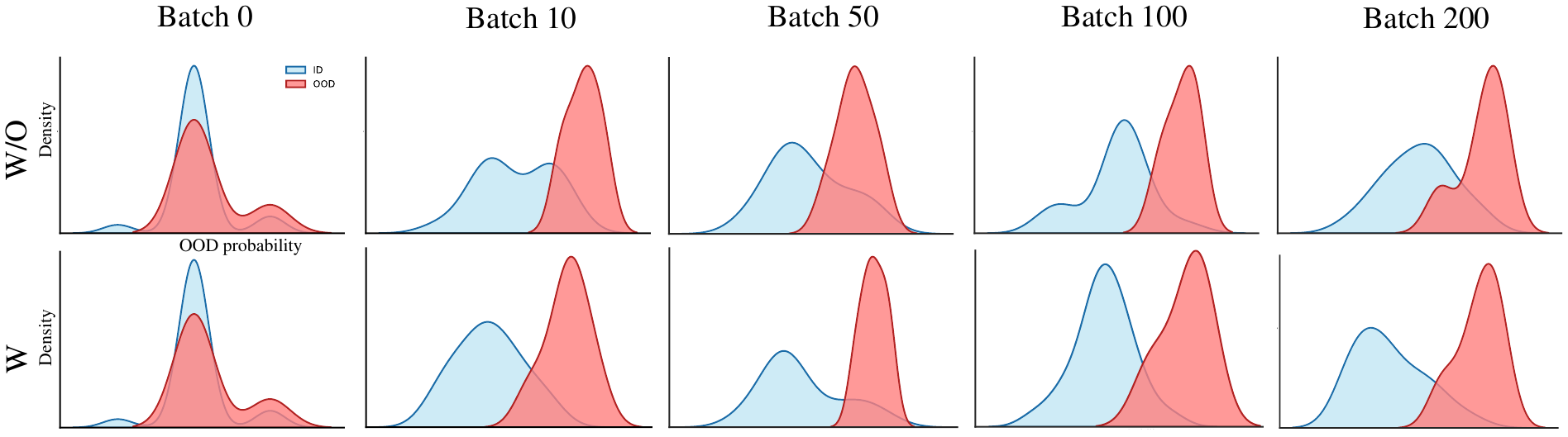}
    \caption{Effect of $\mathcal{L}_{\text{OKP}}$ on separating ID-boundary and OOD samples, 64 samples/batch. Top row: OOD probability density without $\mathcal{L}_{\text{OKP}}$; bottom row: OOD probability density with $\mathcal{L}_{\text{OKP}}$.} 
    \label{fig:L_dist_supple}
\end{figure*}

\begin{table}[htbp] 

         
        \centering
        \captionof{table}{Ablation study of the $L_{OMB}$.}
        \label{tab:l_cbce}
        \begin{tabular}{lcc}
        \toprule
        Function & FPR95$\downarrow$ & AUROC$\uparrow$  \\
        \midrule
         $\mathcal{L}_{\text{CE}}$  & 14.23  & 96.98 \\
         \rowcolor{red!10}
         $\mathcal{L}_{\text{OMB}}$ & \textbf{12.46}  & \textbf{97.29} \\
        \bottomrule
        \end{tabular}
\end{table}

\begin{table*}[htbp]
  \centering
  \caption{Performance Comparison on OpenOOD benchmark across three ID Datasets. Lower FPR95 and higher AUROC are better. Best results are in \textbf{bold}, and the second-best results are \underline{underlined}.}
  \label{tab:ood_performance_openood_without}
  \begin{tabular}{l l c c c c c c}
    \toprule
    \multirow{2}{*}{\textbf{ID Dataset}} & \multirow{2}{*}{\textbf{Method}} & \multirow{2}{*}{\textbf{Extra Resources}} & \multicolumn{2}{c}{\textbf{Near OOD}} & \multicolumn{2}{c}{\textbf{Far OOD}} \\
    \cmidrule(lr){4-5} \cmidrule(lr){6-7}
    & & & \textbf{FPR95 $\downarrow$} & \textbf{AUROC $\uparrow$} & \textbf{FPR95 $\downarrow$} & \textbf{AUROC $\uparrow$} \\
    \midrule
    \multirow{6}{*}{CIFAR 10} 
    & MCM           & $\times$ & 35.00     & 91.00    & 12.57 & 96.77 \\
    & Neglabel      & $\surd$  & 35.32  & 92.96 & 15.74 & 96.29 \\
    & OODD      & $\times$ & 48.61  & 89.19 & 14.11 & 96.70 \\
    & AdaND         & $\times$ & 35.06  & 90.97 & ~\underline{2.02}  & \underline{99.40} \\
    & AdaNeg        & $\surd$  & \underline{32.38} & \textbf{94.01} & ~~7.31  & 98.28 \\
    & \cellcolor{red!10}TTL & \cellcolor{red!10}$\times$ & \cellcolor{red!10}\textbf{30.25} & \cellcolor{red!10}\underline{93.60} & \cellcolor{red!10}\textbf{~~1.07} & \cellcolor{red!10}\textbf{99.75} \\
    \midrule
    \multirow{6}{*}{CIFAR 100} 
    & MCM           & $\times$ & 91.01  & 70.53 & 73.27 & 79.66 \\
    & Neglabel      & $\surd$  & 77.54  & 71.90  & 59.66 & 79.84 \\
    & OODD      & $\times$ & 91.52  & 70.23 & 69.48 & 79.88 \\
    & AdaND         & $\times$ & 78.45  & 70.27  & \underline{22.33} & \underline{90.90} \\
    & AdaNeg        & $\surd$  & \underline{71.62} & \underline{77.56} & 40.81 & 88.41 \\
    & \cellcolor{red!10}TTL & \cellcolor{red!10}$\times$ & \cellcolor{red!10}\textbf{52.34} & \cellcolor{red!10}\textbf{82.33} & \cellcolor{red!10}\textbf{~~0.21} & \cellcolor{red!10}\textbf{99.64} \\
    \midrule
    \multirow{6}{*}{ImageNet-200} 
    & MCM           & $\times$ & 63.66  & 83.66 & 17.97 & 96.13 \\
    & Neglabel      & $\surd$  & 49.83  & 87.61 & \underline{9.36}  & 97.87 \\
    & OODD      & $\times$ & 48.39  & 83.82 & 14.82 & 96.13 \\
    & AdaND         & $\times$ & 53.77  & 83.76 & 14.85 & 96.00 \\
    & AdaNeg        & $\surd$  & \textbf{41.48} & \textbf{88.76} & ~~9.79  & \underline{98.05} \\
    & \cellcolor{red!10}TTL & \cellcolor{red!10}$\times$ & \cellcolor{red!10}\underline{48.16} & \cellcolor{red!10}\underline{87.75} & \cellcolor{red!10}\textbf{~~7.89} & \cellcolor{red!10}\textbf{98.50} \\
    \bottomrule
  \end{tabular}
\end{table*}
\begin{table*}[htbp]
  \centering
  \caption{Performance Comparison on OpenOOD benchmark with ImageNet-1k as ID dataset. Lower FPR95 and higher AUROC are better. Best results are in \textbf{bold}, and the second-best results are \underline{underlined}.}
  \label{tab:imagenet1k_ood}
  \begin{tabular}{l l c c c c c}
    \toprule
    \multirow{2}{*}{\textbf{ID Dataset}} & \multirow{2}{*}{\textbf{Method}} & \multirow{2}{*}{\textbf{Extra Resources}} & \multicolumn{2}{c}{\textbf{Near OOD}} & \multicolumn{2}{c}{\textbf{Far OOD}} \\
    \cmidrule(lr){4-5} \cmidrule(lr){6-7}
    & & & \textbf{FPR95 $\downarrow$} & \textbf{AUROC $\uparrow$} & \textbf{FPR95 $\downarrow$} & \textbf{AUROC $\uparrow$} \\
    \midrule
    \multirow{12}{*}{ImageNet-1k} 
    & MCM           & $\times$ & 84.17  & 69.22 & 44.39 & 90.61 \\
    & Neglabel      & $\surd$  & 69.27 & 75.38 & 23.23 & 94.94 \\
    & LoCoOp        & $\times$ & 82.51  & 68.03 & 33.42 & 92.12 \\
    & IDLike        & $\times$ & 86.23  & 59.51 & 36.11 & 92.16 \\
    & LocalPrompt   & $\times$ & 77.91  & 73.44 & 28.6  & 93.71 \\
    & FA            & $\times$ & 69.21  & \underline{77.97} & 25.78 & 93.67 \\
    \cmidrule(lr){2-7}
    & OODD          & $\times$ & 73.83  & 67.08 & 24.81 & 91.87 \\
    & OODD + FA       & $\times$ & \underline{60.06} & 77.31 & 23.52 & 92.21 \\
    & AdaND         & $\times$ & 80.78 & 68.01 & 27.76 & 92.19 \\
    & AdaND + FA      & $\times$ & 75.40  & 74.28 & 23.05 & 93.53 \\
    & AdaNeg        & $\surd$  & 67.35  & 76.01 & 20.9  & 95.44 \\
    & AdaNeg + FA   & $\surd$  & 76.15  & 73.79 & 19.75 & 95.38\\
    & \cellcolor{red!10}TTL & \cellcolor{red!10}$\times$ & \cellcolor{red!10}89.44 & \cellcolor{red!10}56.3  & \cellcolor{red!10}\underline{19.88} & \cellcolor{red!10}\underline{95.59} \\
    & \cellcolor{red!10}Ours + FA & \cellcolor{red!10}$\times$ & \cellcolor{red!10}\textbf{56.53} & \cellcolor{red!10}\textbf{83.33} & \cellcolor{red!10}\textbf{13.62} & \cellcolor{red!10}\textbf{97.05} \\
    \bottomrule
  \end{tabular}
\end{table*}

\begin{table}[htbp] 
        \vspace{-6pt}
        \centering
        \captionof{table}{Effect of TTL with base detectors of different quality.}
        \vspace{-6pt}
        \resizebox{\linewidth}{!}{
        \label{tab:boundary}
        \begin{tabular}{l cc | l  cc}
        \toprule
        \textbf{Method} & \textbf{FPR95$\downarrow$} & \textbf{AUROC$\uparrow$} & \textbf{Method} & \textbf{FPR95$\downarrow$} & \textbf{AUROC$\uparrow$}  \\
        \midrule
        \rowcolor{gray!20}
         MCM  & 42.77  & 90.76 & MaxLogit  & 67.13  & 83.92 \\
         + \textbf{Ours}  & \textbf{12.46}  & \textbf{97.29} & + \textbf{Ours}  & \textbf{33.34}  & \textbf{91.52} \\
         \midrule
         \rowcolor{gray!20}
          MCM-Var  & 77.60 & 73.80  & MCM-Entropy  & \textbf{84.84} & \textbf{68.64}  \\
         \textbf{Ours}  & \textbf{75.74} & \textbf{74.29} & + \textbf{Ours}  & 85.19 & 58.6   \\
        \bottomrule
        \end{tabular}
}
\vspace{-6pt}
\end{table}

\begin{table*}[htbp]
    \centering
    \setlength{\tabcolsep}{0.5pt}
    \scriptsize
    \caption{The nearest words for each of the 4 context tokens discovered on different OOD datasets. Each row corresponds to one context vector, and the ten nearest words are listed in order of similarity. Words that are in \textbf{bold} indicate those directly related to the OOD dataset’s domain semantics. N/A indicates non-Latin tokens. }
    \begin{tabular}{l l l llllllllll}
    \toprule
        dataset & entry & token & 1 & 2 & 3 & 4 & 5 & 6 & 7 & 8 & 9 & 10 \\ \midrule
        \multirow{4}{*}{None} & \multirow{4}{*}{1} & 1 & a & an & 0 & 1 & 6 & 7 & 5 & 4 & 2 & 3 \\
        ~ & ~ & 2 & photo & photos & pic & 7 & 3 & 2 & 4 & 5 & 0 & 1 \\ 
        ~ & ~ & 3 & of & 1 & 3 & 6 & 7 & 0 & 2 & 5 & 8 & 4 \\ 
        ~ & ~ & 4 & a & an & 0 & 1 & 6 & 7 & 5 & 4 & 2 & 3 \\ \midrule
        \multirow{8}{*}{iNaturalist} & \multirow{4}{*}{1} & 1 & femme & \textbf{plants} & charms & \textbf{nutrients} & \textbf{wildflowers} & adilla & investments & things & resu & teammates \\ 
        ~ & ~ & 2 & selfies & alpaca & alpac & !' & N/A & affirm & dancers & N/A & footage & pictures \\ 
        ~ & ~ & 3 & of & opposite & N/A & behalf & whos & past & N/A & keyword & illustrious & N/A \\ 
        ~ & ~ & 4 & behindthe & bankno & chuckle & playo & residente & N/A & besides & expla & caled & starwar \\ \cmidrule(lr){2-13}
        ~ & \multirow{4}{*}{2} & 1 & \textbf{arugula} & adilla & charms & teammates & femme & dementi & hearts & teammate & cosplayer & \textbf{wildflowers} \\ 
        ~ & ~ & 2 & alpaca & airdrop & alpac & extras & N/A & supportindiefilm & confirms & broadcasts & whilst & wholesale \\ 
        ~ & ~ & 3 & N/A & N/A & N/A & N/A & N/A & .!! & mediation & N/A & past & ++ \\ 
        ~ & ~ & 4 & chuckle & N/A & nrl & behindthe & bankno & playo & condomin & whist & \textbf{fore} & isai \\ \midrule
        \multirow{8}{*}{SUN} & \multirow{4}{*}{1} & 1 & mustache & \textbf{stay} & a & attire & braces & \textbf{cruise} & signature & autograph & scones & \textbf{azure} \\ 
        ~ & ~ & 2 & photo & \textbf{campsite} & \textbf{landscape} & \textbf{seascape} & activation & \textbf{chillout} & \textbf{landscapes} & letstalk & !' & \textbf{landscapephotography} \\ 
        ~ & ~ & 3 & of & \textbf{enjoying} & \textbf{enjoys} & actions & aton & \textbf{enjoyed} & installing & activation & calling & \textbf{exploring} \\ 
        ~ & ~ & 4 & weal & \textbf{evapor} & metaph & qur & poon & mpg & meand & syl & refr & \textbf{wooden} \\ \cmidrule(lr){2-13}
        ~ & \multirow{4}{*}{2} & 1 & mustache & piday & commissioning & autograph & \textbf{stay} & date & \textbf{cruise} & attire & signature & \textbf{typography} \\ 
        ~ & ~ & 2 & photo & \textbf{campsite} & offseason & lineups & gameplay & commitment & \textbf{naturephotography} & \textbf{commuting} & \textbf{chillout} & letstalk \\ 
        ~ & ~ & 3 & \textbf{enjoys} & \textbf{enjoying} & \textbf{exploring} & \textbf{enjoyed} & activation & installing & applied & medallion & \textbf{explores} & \textbf{enjoy} \\ 
        ~ & ~ & 4 & syl & qur & earn & wfa & metaph & ellu & edou & prosp & margare & benevol \\ \midrule
        \multirow{8}{*}{places} & 1 & 1 & a & \textbf{roma} & \textbf{footprints} & N/A & forza & N/A & an & N/A & iftar & etres \\ 
        ~ & ~ & 2 & photo & -' & addresses & painting & [@ & N/A & addressing & semester & \textbf{photobook} & (@ \\ 
        ~ & ~ & 3 & \textbf{seine} & movements & sem & functionality & optimize & \textbf{homeowner} & adn & \textbf{morn} & prompted & of \\ 
        ~ & ~ & 4 & unex & \textbf{terracotta} & unemployment & invent & economists & jailbreak & econ & exclu & dispro & ameli \\ \cmidrule(lr){2-13}\cmidrule(lr){2-13}
        ~ & 2 & 1 & \textbf{ulster} & \textbf{slovenia} & \textbf{cryst} & \textbf{oceans} & N/A & outline & \textbf{surrealism} & rgv & suffra & tracklist \\ 
        ~ & ~ & 2 & hdr & | & N/A & redesigned & installations & N/A & printable & ': & || & \textbf{styled} \\ 
        ~ & ~ & 3 & wkend & \textbf{contempor} & \textbf{morn} & beginner & lovehim & arrog & seine & consult & wks & societal \\ 
        ~ & ~ & 4 & unex & bluebird & spie & dispro & \textbf{statu} & baff & pinst & economists & jailbreak & econ \\ \midrule
        \multirow{8}{*}{Texture} & 1 & 1 & xboxone & N/A & N/A & N/A & N/A & entirety & gianni & trojans & hardware & gmail \\ 
        ~ & ~ & 2 & olympians & ballet & paralympic & sawards & warhol & loool & N/A & fineartamerica & soviet & N/A \\ 
        ~ & ~ & 3 & \textbf{kaleido} & \textbf{formed} & \textbf{abstractart} & rof & elector & atomic & houghton & appointed & sitcom & -\$ \\ 
    ~ & ~ & 4 & exem & womancrush & \textbf{homeitems} & \textbf{viny} & \textbf{plicity} & \textbf{refurbi} & \textbf{ergon} & \textbf{flavo} & rahulg & fianc \\ \cmidrule(lr){2-13}
        ~ & 2 & 1 & a & N/A & every & etres & accessing & momento & ramapho & equip & hahahha & websites \\ 
        ~ & ~ & 2 & photo & \textbf{uniforms} & twitart & N/A & followparty & N/A & mortgages & railwayana & arrested & N/A \\ 
        ~ & ~ & 3 & of & on & from & to & and & of & about & with & in & \& \\ 
        ~ & ~ & 4 & N/A & recap & ksu & N/A & N/A & ksa & \textbf{sculpt} & \textbf{pendants} & dessert & \textbf{nailart} \\ 
        \bottomrule
    \end{tabular}
    \label{tab:token_results_multirow}
\end{table*}

\noindent\textbf{Designs of scoring function with OOD prompts.}
Given the learned OOD prompts, 
Four strategies for generating final calibrated prediction $S_ {final}$ using discovered OOD knowledge are evaluated on ImageNet-1k benchmark: 
(1) Max OOD Similarity (MaxSim);
(2) Exponentiated Sum OOD (ExpSum); and
(3) ID-OOD Softmax Ratio (IDR)~\citep{neglabel}; 
As shown in Table~\ref{tab:ood_knowledge_stra_all}, 
the most effective strategy is ours, which directly subtracts OKB similarity from the base detector score.
This show that OOD knowledge functions as a complementary 
corrective signal, with effectiveness depending primarily on knowledge quality rather than sophisticated scoring mechanisms. 



\noindent\textbf{Different Backbone Architectures.} Please refer to Table~\ref{tab:different_arch}. As shown in the results, TTOD still achieved the best performance under different backbones.

\noindent\textbf{Ordering of Testing Data.} Test-time adaptation methods are inevitably affected by the order in which the test samples arrive. To rigorously test this aspect, we randomly shuffled the order of the test data using three distinct seeds. We observed that our method exhibits robustness to changes in the ordering of test data. Specifically, across three experiments conducted on the ImageNet dataset, the AUROC scores were 97.34\%, 97.2\%, and 97.34\%, respectively, demonstrating fluctuations of less than 0.2\%. We report the average results from three random runs in our paper.

\noindent\textbf{Effect of $\mathcal {L}_ {\text{OKP}}$.} 
 As shown in Figure \ref{fig:L_dist_supple}, with $\mathcal{L}_{\text{OKP}}$, the collected pseudo-labeled OOD set progressively splits into a much clearer bimodal distribution over time.
This confirms that $\mathcal{L}_{\text{OKP}}$ successfully separates OOD samples from ID boundary samples, the source of contamination, ultimately enabling more reliable test-time adaptation for OOD detection.

\noindent\textbf{Variants of $\mathcal{L}_{\text{OMB}}$.} $\mathcal{L}_{\text{OMB}}$  consistently outperforms standard cross-entropy $\mathcal{L}_{\text{CE}}$ (Table \ref{tab:l_cbce}) on ImageNet-1k benchmark. By balancing the weight of ID/OOD sample weights, $\mathcal{L}_{\text{OMB}}$ 
effectively preserving the learning signal of minority-class samples. 
This yields more balanced updates and improved OOD detection robustness.



\noindent\textbf{Effectiveness on OpenOOD Benchmark.} We use four popular ID datasets CIFAR-10/100~\citep{cifar}, ImageNet-200/1K~\citep{DBLP:conf/cvpr/DengDSLL009}. Following the OpenOOD benchmark~\citep{openood}, the OOD testing datasets are categorized into two groups: Near OOD and Far OOD. Specifically, for CIFAR-10/100 benchmarks, the Far OOD group includes MNIST~\citep{mnist}, SVHN~\citep{SVHN}, Texture~\citep{DBLP:conf/cvpr/CimpoiMKMV14}, Places365~\citep{DBLP:journals/pami/ZhouLKO018}, and the Near OOD group comprises CIFAR-100/10 and Tiny ImageNet-200~\citep{DBLP:conf/cvpr/DengDSLL009}. For ImageNet-200/1K, the Near OOD group includes SSB-hard~\citep{ssbhard}, NINCO~\citep{ninco}, and the Far OOD group comprises iNaturalist~\citep{DBLP:conf/cvpr/HornASCSSAPB18}, Texture~\citep{DBLP:conf/cvpr/CimpoiMKMV14}, and OpenImage-O~\citep{openimage-o}.

The base detectors employed by OODD, AdaND, and our method TTL are all MCM, while AdaNeg utilizes the more powerful Neglabel as its base detector.
As shown in Table~\ref{tab:ood_performance_openood_without} and Table~\ref{tab:imagenet1k_ood} on the OpenOOD benchmark~\citep{openood}, under the far-out-of-distribution (far-OOD) setting, our method consistently achieves the best results. Under the near-OOD setting, when using CIFAR-10, CIFAR-100, and ImageNet-200 as in-distribution (ID) datasets, our method's performance rivals that of AdaNeg using external datasets and stronger base detectors. This demonstrates the effectiveness of our method across diverse experimental settings. 

In the most challenging near OOD experimental setting using ImageNet-1k as the ID dataset, all methods exhibit limited performance.
Since the base detector is fundamentally incapable of effectively distinguishing between ID and OOD samples, the OODD, AdaND, and TTL methods—which employ weaker base detectors—failed to achieve effective adaptation. Their performance was, to varying degrees, inferior to that of the base detector.
Conversely, AdaNeg, utilizing a stronger base detector, still achieves performance improvements. For a fairer comparison, we uniformly replaced the base detectors of all adaptation methods with the stronger FA detector for testing. Results indicate that when the base detector possesses sufficient discrimination capability, our method achieves the best adaptation performance.




\noindent\textbf{Analysis of Learned OOD Textual Knowledge.} To interpret the textual knowledge learned and stored in the OKB during test-time OOD detection, we draw inspiration from CoOp’s analytical framework~\citep{coop}. Specifically, we search the vocabulary of the VLM for tokens whose embeddings are closest to the learned vector, based on Euclidean distance. Note that CLIP~\citep{clip} uses the BPE~\citep{bpe} representation for tokenization, and thus its vocabulary contains frequent subword units, such as "fore” (shared by words like “forest” and "foreland”).  In our experiments, the learnable prompt consists of four tokens. Since these prompts vectors are optimized in a continuous space, we visualize their semantic meanings by listing the ten nearest words to each learned vector. The Table \ref{tab:token_results_multirow} shows results on the ImageNet-1k benchmark. The “None” entry indicates the default prompt initialized with “a photo of a”, while two representative entries from the OKB are shown for illustration.
On the \textit{iNaturalist} dataset, the nearest tokens such as \textit{plants}, \textit{wildflowers}, \textit{alpaca}, and \textit{arugula} demonstrate that the model autonomously learns ecological and biological semantics from unseen natural images. 
On the \textit{SUN} dataset, dominated by diverse outdoor scenes, the discovered tokens—\textit{landscape}, \textit{seascape}, \textit{campsite}, and \textit{cruise}—indicate that the model captures spatial composition and environmental context as OOD semantic cues. 
The emergence of verbs such as \textit{enjoying}, \textit{exploring}, and \textit{commuting} further suggests that the model identifies human–scene interaction semantics absent from the ID training domain, demonstrating its capacity to extract high-level contextual shifts for better detection. 
In the \textit{Places} dataset, which represents complex scene-level OOD distributions, the discovered tokens (e.g., \textit{footprints}, \textit{oceans}, \textit{terracotta}, \textit{surrealism}) reflect both material and stylistic semantics that differ from the in-distribution visual domain. By learning these distinctive scene-level attributes without supervision, the model effectively captures the semantic shift between familiar and unseen environments. 
For the texture-dominant \textit{Texture} dataset, the nearest tokens—\textit{formed}, \textit{abstractart}, \textit{homeitems}, \textit{refurbi}, and \textit{flavo}—show that the model adapts to surface- and pattern-related semantics. 
These learned tokens correspond to appearance-level variations such as color, gloss, and regularity, indicating that the model learns texture-sensitive knowledge that enhances its ability to detect OOD samples exhibiting distributional drift at the visual level. 
Overall, these results demonstrate that even without explicit retraining, our method can \textbf{discover OOD-relevant textual knowledge} from unlabeled test-time data streams. 
The learned semantics capture and describe distributional discrepancies between ID and OOD domains in an interpretable way, thereby improving both the model’s adaptability and the robustness of OOD detection under distribution shift.

\noindent\textbf{Robustness to Base Detector Quality.}
Pseudo-label noise and error reinforcement are an inherent challenge in test-time adaptation, yet our evaluation across varying signal-to-noise ratios (Table~\ref{tab:boundary}) demonstrates that TTL acts as a robust signal amplifier rather than relying on a strong oracle. Specifically, TTL yields consistent gains even when the base detector is weak (e.g., MCM-Var~\citep{mcm}), confirming that our purification mechanism successfully extracts structural information from noisy pseudo-labels. It is only when the base signal's signal approaches a near-random regime (e.g., MCM-Entropy~\citep{mcm}) that adaptation naturally diminishes. These results confirm that TTL does not require high-quality supervision but effectively improves performance provided the base detector offers a weak-but-structured separation signal.

\end{document}